\documentclass{article} %
\usepackage{iclr2027_conference,times}

\newif\ifarxiv
\arxivtrue
\ifarxiv\iclrfinalcopy\fi

\usepackage[T1]{fontenc}
\usepackage[dvipsnames]{xcolor}  %
\usepackage{amsmath}
\usepackage{amssymb}             %
\usepackage[hyphens]{url}

\usepackage{eccvabbrv}  %
\usepackage{graphicx}
\usepackage{booktabs}
\usepackage[accsupp]{axessibility}

\usepackage[utf8]{inputenc}
\usepackage{amsfonts}
\usepackage{nicefrac}
\usepackage{microtype}
\usepackage{bm}
\usepackage{multirow}
\usepackage{enumitem}  %
\usepackage{tikz}
\usetikzlibrary{arrows.meta, decorations.pathreplacing, shapes.geometric, shadows}
\usepackage{makecell}
\usepackage{colortbl}
\usepackage{siunitx}
\usepackage{xspace}

\definecolor{eccvblue}{rgb}{0.12,0.49,0.85}
\usepackage[pagebackref,breaklinks,colorlinks,bookmarks=false,urlcolor=eccvblue,citecolor=eccvblue]{hyperref}

\usepackage{orcidlink}

\usepackage[labelfont=bf,font=small]{caption}
\usepackage[skip=3pt]{subcaption}
\usepackage[capitalize]{cleveref}  %

\newcommand{\ours}{ZipSplat}

\newcommand{\comment}[1]{}

\definecolor{tabfirst}{HTML}{adc6ae}   %
\definecolor{tabsecond}{HTML}{efe3ad}  %
\definecolor{tabthird}{HTML}{e3edf9}   %
\newcommand{\cfirst}{\cellcolor{tabfirst}\bfseries}
\newcommand{\csecond}{\cellcolor{tabsecond}}
\newcommand{\cthird}{\cellcolor{tabthird}}

\renewcommand{\*}[1]{\bm{\mathrm{#1}}}

\newcommand{\real}{\mathbb{R}}

\newcommand{\loss}{\mathcal{L}}

\newcommand{\nviews}{N}              %
\newcommand{\ngs}{G}                 %
\newcommand{\qr}{r}                  %
\newcommand{\softplus}{\operatorname{softplus}}

\newcommand{\gmean}{\boldsymbol{\mu}}  %
\newcommand{\gscale}{\*{s}}            %
\newcommand{\grot}{\*{q}}              %
\newcommand{\gopac}{\alpha}            %
\newcommand{\gcolor}{\*{c}}            %

\newcommand{\vtok}{\*{f}}              %
\newcommand{\gtok}{\*{g}}              %
\newcommand{\stok}{\*{z}}              %
\newcommand{\stokset}{\*{Z}}           %

\newcommand{\vtoknames}{visual tokens\xspace}
\newcommand{\stokname}{scene token\xspace}
\newcommand{\stoknames}{scene tokens\xspace}
\newcommand{\pose}{\*{P}}              %

\newcommand{\spoint}{\*{p}}            %
\renewcommand{\vs}{\textit{vs.}\xspace}

\renewcommand{\paragraph}[1]{\par\noindent\textbf{#1}}

\newif\ifaddsupp
\addsupptrue

\ifarxiv
\title{\ours: Fewer Gaussians, Better Splats}
\else
\title{\ours: Observation-Grounded Scene Tokens for Feed-Forward Gaussian Splatting}
\fi

\author{Alexander Veicht$^{1}$, Sunghwan Hong\thanks{Corresponding author}~$^{1}$, D\'aniel Bar\'ath$^{1}$ \& Marc Pollefeys$^{1,2}$ \\
$^{1}$ETH Z\"urich \quad $^{2}$Microsoft}

\begin{document}

\maketitle
\ifarxiv\lhead{Preprint}\fi
\noindent\hspace*{\fill}\begin{minipage}{0.99\linewidth}
\centering
\includegraphics[width=\linewidth]{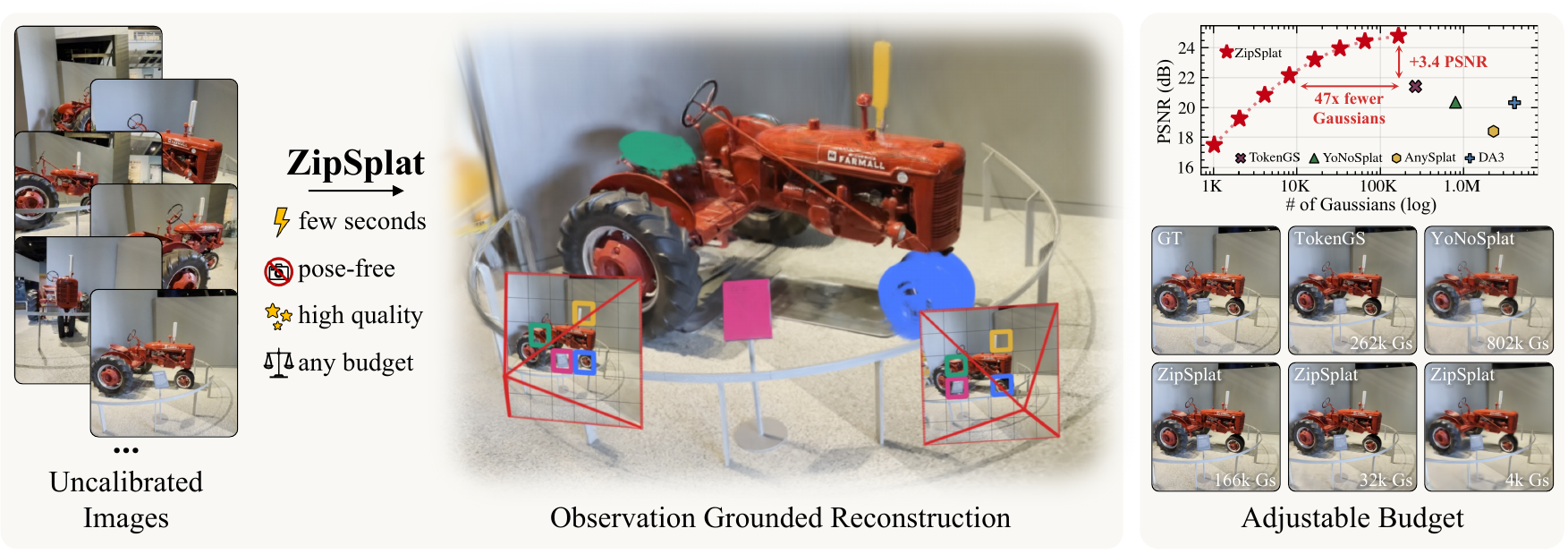}
\captionof{figure}{
    \textbf{\ours{} reconstructs scenes from unposed images in seconds.}
    Each scene token is derived from a source image patch and decoded into freely placed Gaussians that are grounded in their observation  (colored groups $\leftrightarrow$ patches).
    Redundant tokens across views are merged into a compact set whose size is adjustable at inference without retraining.
    \textit{Right:} the same DL3DV scene from 16 views at 166K, 32K, and 4K Gaussians.
    \ours{} matches the strongest baseline with $47{\times}$ fewer Gaussians and surpasses it by $3.4$\,dB at full budget.
}
\label{fig:teaser}
\end{minipage}\hspace*{\fill}

\ifarxiv\vspace{9pt}\fi

\begin{abstract}
Feed-forward 3D Gaussian Splatting reconstructs scenes from images in seconds. 
Pixel-aligned methods keep every Gaussian grounded in an observation, but produce redundant, potentially misaligned Gaussians across views. 
Token-based methods instead aggregate all views into a fixed set of learned tokens but lack explicit observation support and use a fixed reconstruction budget.
We introduce \ours{}, a method that predicts a compact, observation-grounded scene representation whose size is set at inference.
It derives each scene token from an image patch rather than a learned query, and decodes it into freely placed 3D Gaussians.
Tokens across views are merged into a compact set whose size is adjustable at inference.
However, without geometric supervision, decoded Gaussians can drift from their source patch, and merging can mix tokens from different surfaces.
We therefore introduce two geometric losses to ground and shape this representation.
A cone loss ties each token's Gaussians to its source patch, and GeoCE aligns token similarity with 3D proximity so that tokens observing the same surface merge together.
Building on these observation--geometry associations, we further introduce TokenBA, a token-level bundle adjustment that refines scene tokens and camera poses in a few seconds.
\ours{} requires neither camera poses nor intrinsics, yet outperforms pixel-aligned and token-based baselines on DL3DV and RealEstate10K, even those given ground-truth cameras.
It further generalizes zero-shot to Mip-NeRF360 and ScanNet++.
\ificlrfinal\else Code and trained models will be made public.\fi
\ificlrfinal Our project page is at {\href{https://veichta.com/zipsplat}{https://veichta.com/zipsplat}}.\fi
\end{abstract}

\section{Introduction}
\label{sec:intro}

Novel view synthesis has shifted from per-scene optimization of radiance fields~\citep{mildenhall2020nerf} and 3D Gaussians~\citep{kerbl20233d} to feed-forward models that predict a Gaussian scene from a collection of images in a few seconds~\citep{charatan2024pixelsplat,chen2024mvsplat,xu2025depthsplat,xu2026resplat}. 
As these methods mature, performance increasingly depends on the scene representation itself: how image observations become Gaussians, how many are predicted, and where they are placed.

Most existing feed-forward methods predict one Gaussian per input pixel~\citep{xu2025depthsplat,ye2024no,ye2026yonosplat,jiang2025anysplat}, directly grounding each prediction in an image observation.
However, repeated observations of the same surface produce separate, view-specific Gaussians rather than a shared prediction, creating substantial cross-view redundancy.
Errors in the depth and camera estimates can further misalign these predictions and produce visible artifacts.

Recent methods based on fixed latent queries instead aggregate multi-view observations into shared tokens before decoding freely placed Gaussians~\citep{an2025c3g,ren2026tokengs,itkin2026globalsplat}.
This aggregation can reduce redundancy and cross-view misalignment, but the learned queries have no explicit link to the observations they represent. 
Unlike pixel-aligned Gaussians, whose source view is known by construction, these tokens must learn their observation support implicitly, which can produce blurry reconstructions.
Their reconstruction budget is also fixed by design: a small object capture and a large scene receive the same number of tokens and Gaussians.

These limitations motivate a representation that grounds each prediction in its source observation without fixing the reconstruction budget in the architecture.
We introduce \textbf{\ours{}}, which predicts a compact, observation-grounded scene representation from unposed images.
Its reconstruction budget is adjustable at inference without retraining, letting it match the scene rather than the architecture.

Rather than using learned queries, we derive each scene token directly from its source image patch, making the token-to-observation link explicit by construction.
Each token is decoded into freely placed 3D Gaussians, but without geometric constraints the decoded Gaussians can drift from the patch that produced them.
A new cone loss addresses this by constraining each token's Gaussians to reproject near their source patch, grounding them in their observation.

As views overlap, many patches observe the same surface and their tokens can be merged into a more compact set.
However, merging in feature space does not guarantee that the resulting groups correspond to the same surface.
A new geometric contrastive loss, GeoCE, aligns token similarity with 3D proximity so that clusters form by surface rather than by viewpoint.

The representation is now compact and grounded, but it remains the output of a single forward pass, with residual camera and geometry error.
The grounding that ties each token's Gaussians to its source patch also gives every token a persistent 2D--3D correspondence that can be used for geometric refinement.
We exploit this with TokenBA, a token-level bundle adjustment that refines cameras and scene tokens through the frozen decoder, improving reconstruction in seconds.

\ours{} establishes a new state of the art on standard benchmarks using only unposed, uncalibrated images at inference.
It consistently outperforms both pixel-aligned and token-based feed-forward methods while using substantially fewer Gaussians.
It generalizes zero-shot to diverse indoor and outdoor scenes, and continues to improve as additional views are provided.
Ablations validate the representation and its components, and further experiments characterize the adjustable budget and cross-dataset generalization.

In summary, we make the following contributions:
\begin{itemize}[leftmargin=1.2em,itemsep=1pt,topsep=2pt,parsep=0pt,label={\scalebox{0.6}{\textbullet}}]
  \item We introduce \ours{}, a feed-forward model that reconstructs scenes from unposed images as compact, observation-grounded \stoknames{}, each derived from a source patch and decoded into freely placed Gaussians. The reconstruction budget is adjustable at inference without retraining.
  
  \item We introduce two geometric losses that ground and shape this representation geometrically: a cone loss that ties each token's decoded Gaussians to its source patch, and GeoCE, which aligns token similarity with 3D proximity for cross-view clustering.
  
  \item We introduce TokenBA, an efficient token-level bundle adjustment that exploits the resulting 2D--3D correspondences to jointly refine \stoknames{} and camera poses through reprojection and photometric consistency within a few seconds.
\end{itemize}

\section{Related Work}
\label{sec:related}

\paragraph{Feed-forward 3D Gaussian Splatting.}
Per-scene optimization of radiance fields~\citep{mildenhall2020nerf,barron2022mipnerf360} and 3D Gaussians~\citep{kerbl20233d} produces high-quality novel views but is costly.
Feed-forward methods instead predict Gaussians in a single pass, using epipolar attention, cost volumes, or scalable transformers~\citep{charatan2024pixelsplat,chen2024mvsplat,xu2025depthsplat,zhang2024gs,ziwen2025long}.
Pose-free variants progressively remove camera assumptions~\citep{ye2024no,smart2024splatt3r,hong2024pf3plat,huang2025no,kang2025selfsplat}, with recent methods jointly estimating cameras and Gaussians from fully uncalibrated inputs~\citep{jiang2025anysplat,zhang2025flare,ye2026yonosplat,lin2025depth}.
Most, however, retain a pixel-to-Gaussian mapping: this grounds each prediction but makes the Gaussian count scale with the input rather than scene complexity.
Pruning, merging, and sparse pixel selection reduce this set only after per-view prediction~\citep{ziwen2025long,jiang2025anysplat,wang2024freesplat,zhang2024gaussian,wang2026volsplat,park2025ecosplat,zhang2026sparsesplat,moreau2026offthegrid,fei2024pixelgaussian,kang2026ilrm}, while input compression~\citep{wang2026zpressor} and storage codecs~\citep{chen2025fcgs,song2026tinysplat,lee2024compact,morgenstern2024compact} do not change this representation.

\paragraph{Token-based Gaussian prediction.}
Token-based methods remove pixel alignment by decoding freely placed Gaussians from latent tokens, similar to Scaffold-GS~\citep{lu2024scaffold}, which decodes them from per-scene anchors.
TokenGS~\citep{ren2026tokengs}, C3G~\citep{an2025c3g}, and GlobalSplat~\citep{itkin2026globalsplat} aggregate multi-view features with learned queries, while TokenSplat~\citep{li2026tokensplat} voxelizes and merges image tokens before decoding.
These approaches fuse repeated observations and decouple the Gaussian count from input size, but their query count or voxelization is fixed, so changing the budget requires retraining.
Learned queries also lack explicit supporting observations.
In contrast, \ours{} clusters image tokens in a learned geometric space, preserving their observation support while allowing the budget to be adjusted at inference through the clustering ratio.

\paragraph{Post-hoc Refinements.}
Learned reconstruction systems commonly end with a refinement step: DROID-SLAM~\citep{teed2021droid} embeds a dense bundle adjustment layer in its update, and VGGT~\citep{wang2025vggt} refines its predicted cameras by bundle adjustment over predicted tracks.
Feed-forward 3DGS methods so far ignore or lack such correspondences, so refinement falls back to photometric optimization with ground-truth poses.
C3G~\citep{an2025c3g} optimizes the decoded Gaussians as in per-scene 3DGS, while TokenGS~\citep{ren2026tokengs} tunes its learnable queries through the frozen network, which preserves geometry but trails Gaussian-space tuning in quality.
TokenBA instead uses the built-in 2D--3D correspondences of \stoknames{} to refine the scene representation and camera poses, achieving faster convergence and higher quality without pose supervision.

\section{Method}
\label{sec:method}
\begin{figure*}[t]
\centering
\includegraphics[width=\linewidth]{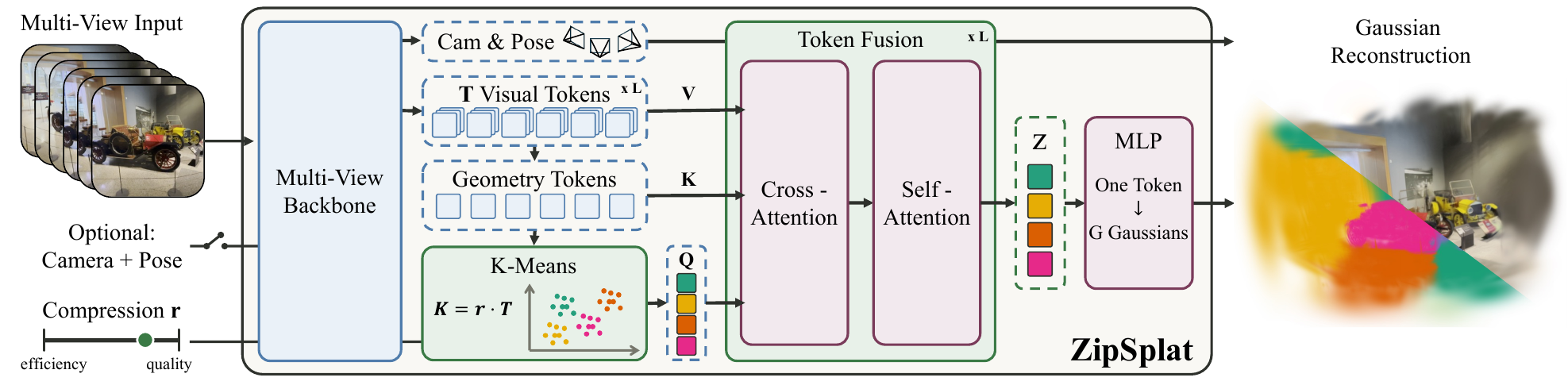}
\caption{
\textbf{Overview of \ours.}
Given $\nviews$ unposed images, a multi-view backbone extracts per-patch geometry tokens that are clustered into $K$ \stoknames{} using $k$-means.
Each \stokname{} is derived from its source patches, and cross- and self-attention refine them.
An MLP finally decodes each into $\ngs$ spatially clustered Gaussians (colors denote tokens).
The number of \stoknames{} $K$ is set by the compression ratio $\qr$ at inference, giving a single model a continuous quality--efficiency trade-off.
}
\label{fig:architecture}
\end{figure*}

Given $\nviews$ unposed and uncalibrated images $I = \{I_n \in \real^{H \times W \times 3}\}_{n=1}^{\nviews}$, \ours{} predicts a set of 3D Gaussians~\citep{kerbl20233d} and estimates the camera parameters.
Each Gaussian is parameterized by a center $\gmean \in \real^3$, scales $\gscale \in \real^3$, rotation quaternion $\grot \in \real^4$, opacity $\gopac \in [0,1]$, and color coefficients $\gcolor \in \real^{C}$.
Novel views are rendered via differentiable rasterization~\citep{ye2025gsplat}.

\paragraph{Overview.}
\Cref{fig:architecture} summarizes the pipeline.
A multi-view backbone produces geometry tokens from per-patch features.
These are clustered into \stoknames{}, refined through cross- and self-attention, and decoded into freely placed 3D Gaussians, while camera parameters are estimated alongside.
We derive each \stokname{} from its source patches, and a cone loss grounds its Gaussians near them, which gives persistent 2D--3D correspondences (\cref{sec:tokens}).
Clustering removes cross-view redundancy at an adjustable budget, guided by GeoCE (\cref{sec:merging}), and TokenBA refines cameras and tokens via the patch--Gaussian associations (\cref{sec:tokenba}).

\subsection{Observation-Grounded Scene Tokens}
\label{sec:tokens}

Each image patch observes a specific surface region.
To give each observation a dedicated token, a multi-view backbone extracts per-patch \vtoknames{} at multiple layers.
These are merged into geometry tokens $\gtok$, one per image patch, forming a shared space for clustering, attention, and geometric supervision.
Each geometry token serves as a \stokname{}.

\paragraph{Gaussian decoding.}
Each \stokname{} represents a surface region whose spatial extent and complexity vary across the scene.
An MLP decoder $D$ maps each \stokname{} $\stok_k$ into $\ngs$ freely placed Gaussians:
\begin{equation}
  \mathcal{G}_k
  =
  \left\{
  \left(
  \gmean_{kj},
  \gscale_{kj},
  \grot_{kj},
  \gopac_{kj},
  \gcolor_{kj}
  \right)
  \right\}_{j=1}^{\ngs}
  =
  D(\stok_k),
  \label{eq:decode}
\end{equation}
with the full scene assembled from all $K$ tokens.
Free placement lets the token spread its Gaussians over the observed region rather than anchoring each one to a fixed ray.
However, nothing constrains decoded Gaussians to stay near their source patch: the decoder can place them anywhere in 3D, and the association between token and observation is not guaranteed.

\begin{figure*}[t]
\centering
\captionsetup[subfigure]{skip=1pt}%
\ifdefined\lossh\else\newlength{\lossh}\fi%
\setlength{\lossh}{0.21\linewidth}%
\begin{subfigure}[b]{0.32\linewidth}
\centering
\includegraphics[height=\lossh]{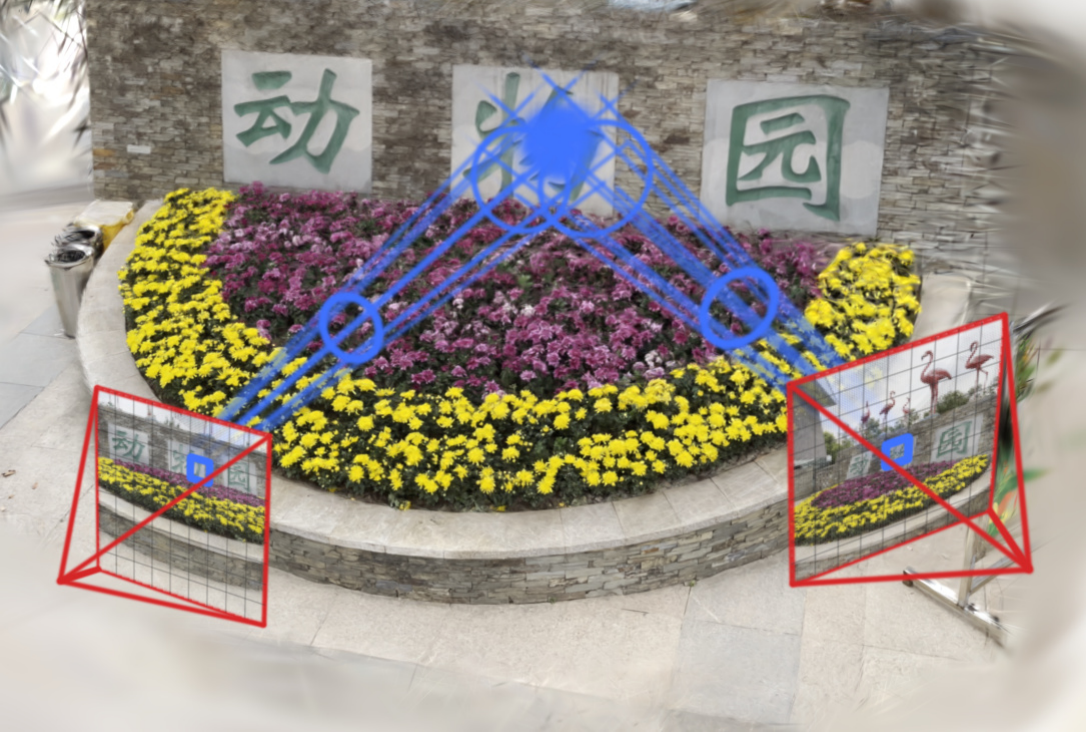}
\caption{Cone Loss}
\label{fig:losses-cone}
\end{subfigure}%
\hfill
\begin{subfigure}[b]{0.32\linewidth}
\centering
\includegraphics[height=\lossh]{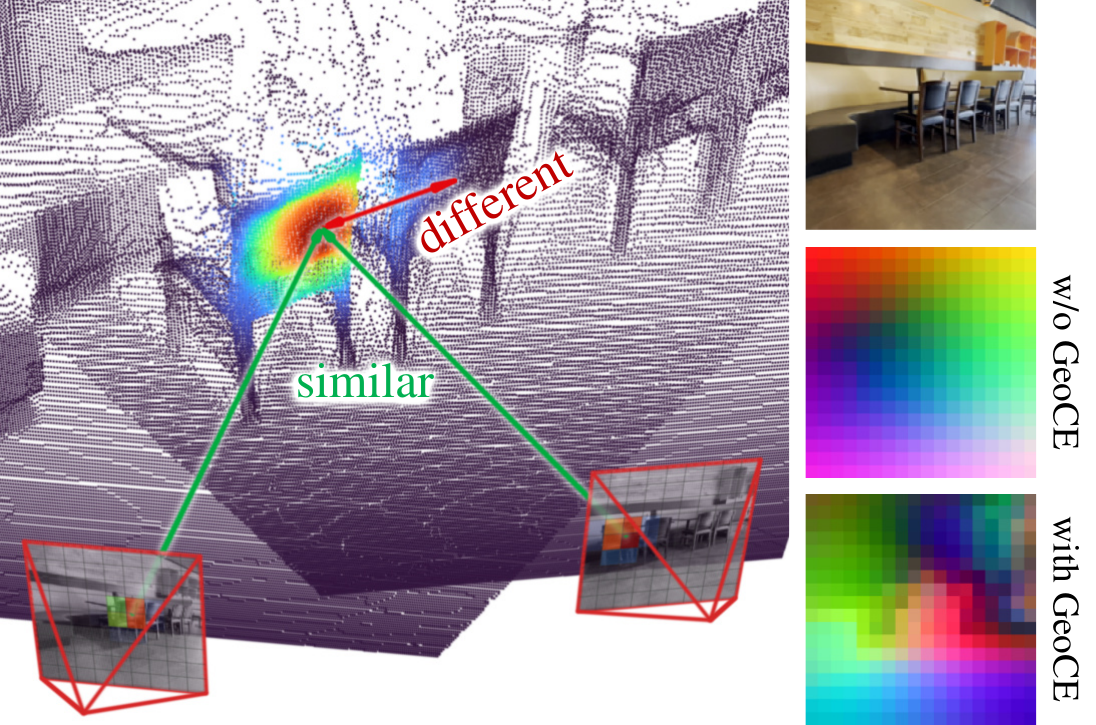}
\caption{GeoCE Loss}
\label{fig:losses-geoce}
\end{subfigure}%
\hfill
\begin{subfigure}[b]{0.32\linewidth}
    \centering
    \includegraphics[height=\lossh]{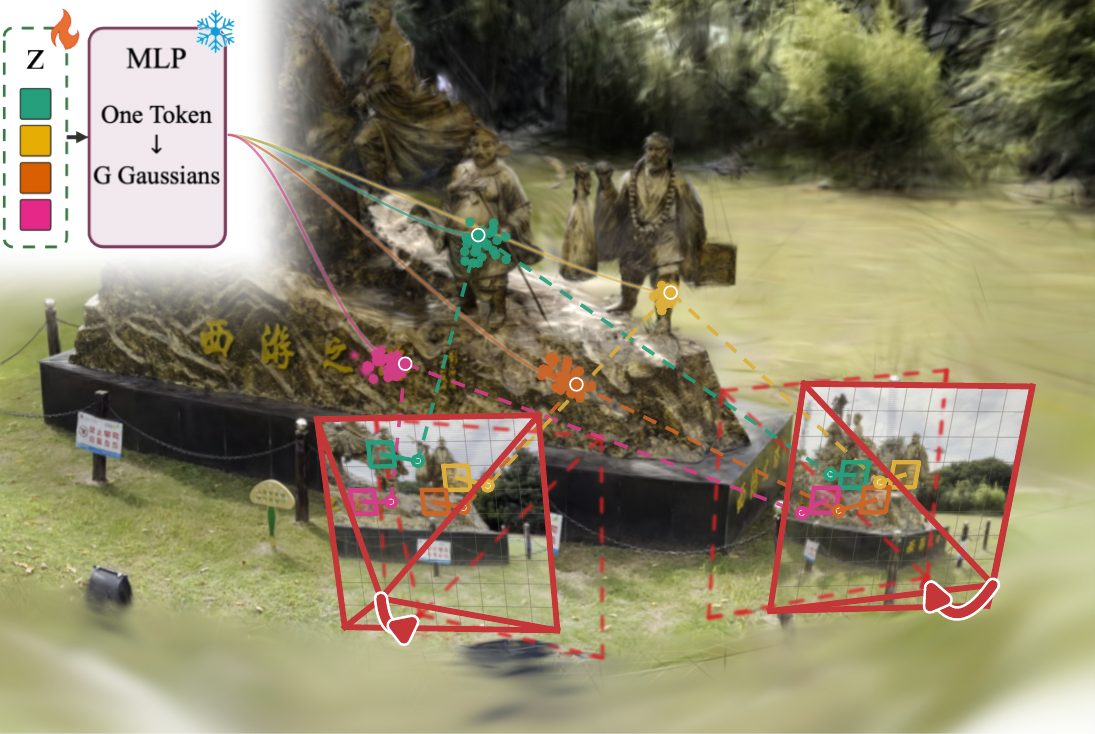}
\caption{TokenBA}
\label{fig:losses-tokenba}
\end{subfigure}
\caption{
\textbf{Grounding, shaping, and refining \stoknames{}.}
(a) The cone loss ties each token's Gaussians back to its source patch, grounding the decoded geometry in the observation that produced it.
(b) GeoCE aligns token similarity with 3D proximity so that clustering merges patches observing the same surface.
(c) TokenBA refines cameras via robust PnP on the patch--Gaussian correspondences and updates scene tokens through the frozen decoder using photometric error on the input views.
}
\label{fig:losses}
\end{figure*}

\paragraph{Cone loss.}
The cone loss grounds each token's Gaussians in its source patch (\cref{fig:losses-cone}).
For each token $k$, it penalizes the reprojection distance between its Gaussian centers and the source patch center:
\begin{equation}
  \loss_{\mathrm{cone}}
  =
  \sum_{k=1}^{K}
  \sum_{j=1}^{\ngs}
  \rho\!\left(
  \left\|
  \pi_k(\gmean_{kj})-\*{u}_k
  \right\|
  \right),
  \label{eq:cone}
\end{equation}
where $\pi_k$ projects into the source view of token $k$, $\*{u}_k$ is its patch center, distances are measured in patch units, and $\rho$ is a robust penalty that limits the influence of large residuals.
This constraint is only definable because each token has a known source patch.
A frustum loss, as used by TokenGS~\citep{ren2026tokengs}, keeps Gaussians visible but cannot tie them to a specific observation.
With the cone loss, our approach yields an observation-grounded scene representation: each \stokname{} is derived from a source patch, its Gaussians are tied back to that patch, and these associations persist at inference.

\subsection{Merging Redundant Scene Tokens}
\label{sec:merging}

When multiple views cover a scene, many patches observe the same surface from different viewpoints.
Their tokens carry redundant information and can be merged into a smaller set whose size is adjustable at inference without retraining.

\begin{figure*}[t]
\centering
\ifdefined\cpw\else\newlength{\cpw}\fi%
\ifdefined\cpairgap\else\newlength{\cpairgap}\fi\setlength{\cpairgap}{5pt}%
\ifdefined\cinnergap\else\newlength{\cinnergap}\fi\setlength{\cinnergap}{2pt}%
\ifdefined\crowlabw\else\newlength{\crowlabw}\fi\setlength{\crowlabw}{0pt}%
\setlength{\cpw}{\dimexpr(\linewidth-\crowlabw-2\cpairgap-3\cinnergap-2pt)/6\relax}%
\newcommand{\cimgv}[1]{\includegraphics[width=\cpw]{figures/compression/assets_v2/#1.pdf}}%
\newcommand{\crow}[1]{%
  \cimgv{r1_#1_rgb_annot}\hspace{\cinnergap}\cimgv{r1_#1_token}\hspace{\cpairgap}%
  \cimgv{r0.2_#1_rgb_annot}\hspace{\cinnergap}\cimgv{r0.2_#1_token}\hspace{\cpairgap}%
  \cimgv{r0.05_#1_rgb_annot}\hspace{\cinnergap}\cimgv{r0.05_#1_token}%
}
\newcommand{\clab}[1]{\makebox[\dimexpr2\cpw+\cinnergap\relax][c]{\small #1}}
\newcommand{\crowlab}[1]{\parbox[b][\cpw][c]{\crowlabw}{\centering\rotatebox{90}{\scriptsize #1}}}
\makebox[\linewidth][c]{\crow{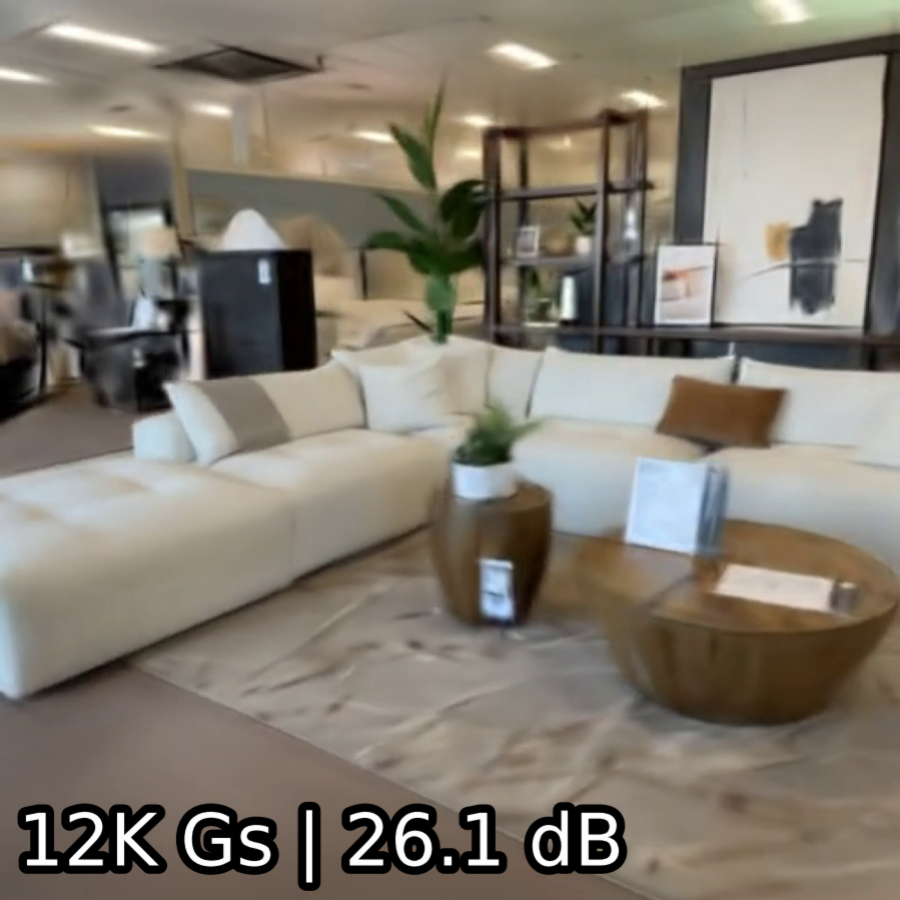}}\par\nointerlineskip\vspace{\cinnergap}%
\makebox[\linewidth][c]{\makebox[\crowlabw]{}\clab{full capacity}\hspace{\cpairgap}\clab{$5\times$ compression}\hspace{\cpairgap}\clab{$20\times$ compression}}
\caption{
\textbf{One model, any budget.}
The same scene at three compression ratios $\qr$, set at inference without retraining. Each pair shows the rendered view (left) and Gaussians colored by their \stokname{} (right).
With more compression, fewer tokens need to cover larger regions but the render stays sharp.
}
\label{fig:compression}
\end{figure*}

\paragraph{Inference-time clustering.}
We cluster the geometry tokens into $K$ groups using $k$-means in cosine space.
Given $T$ patches per view, the budget $K$ is set through the compression ratio
\begin{equation}
  \qr = \frac{K}{\nviews T},
  \qquad \qr\in(0,1].
  \label{eq:compression_ratio}
\end{equation}
Cluster initialization favors photometrically complex regions to preserve fine detail.
Each \stokname{} is initialized as the mean of its cluster members, $\stok_k^{0}=|C_k|^{-1}\sum_{i\in C_k}\gtok_i$, and retains them as supporting observations.
Assignments remain discrete, while gradients propagate through the cluster means during training.
A merged \stokname{} inherits the patches of all its members, and the cone loss takes the minimum over them, $\min_{i\in C_k}\|\pi_i(\gmean_{kj})-\*{u}_i\|$, so each Gaussian must fall into the view cone of at least one member patch.
The free region is the union of the member cones and grows with the cluster.
\Cref{fig:compression} shows the effect of varying $\qr$.

\paragraph{Token fusion.}
To recover detail lost during cluster averaging, we refine the \stoknames{} through alternating cross- and self-attention:
\begin{equation}
  \stok^{l}
  =
  \operatorname{SA}\!\left(
    \operatorname{CA}
    (\stok^{l-1},\,\gtok,\,\vtok_l)
  \right),
  \qquad l=1,\ldots,L,
  \label{eq:fusion}
\end{equation}
where the geometry tokens $\gtok$ serve as keys, biasing retrieval toward relevant observations across views, and the backbone features $\vtok_l$ provide the values, processed from deep to shallow layers.
A final residual cross-attention incorporates color features:
\begin{equation}
  \stok
  =
  \stok^{L}
  +
  \operatorname{CA}
  (\stok^{L},\,\gtok,\,\vtok^{\mathrm{col}}).
  \label{eq:color_ca}
\end{equation}

\paragraph{GeoCE loss.}
Clustering can merge tokens across views, but nothing guarantees it groups patches that observe the same surface.
GeoCE makes this explicit by shaping the token space (\cref{fig:losses-geoce}).
During training, each token's patch is unprojected to a 3D anchor $\spoint_i$ using its median depth.
From these anchors we construct a soft target distribution over token pairs, weighted by 3D proximity:
\begin{equation}
  w_{ij}
  \propto
  \exp\!\left(
    -\frac{\|\spoint_i-\spoint_j\|^2}
           {2\sigma_i\sigma_j}
  \right),
  \qquad j\neq i,
  \label{eq:geoce_target}
\end{equation}
where $\sigma_i$ scales with the local patch spacing (\cref{sec:supp-geoce}).
Nearby patches get high weight; distant ones get near zero.
GeoCE pushes the token cosine similarities to match this target via cross-entropy:
\begin{equation}
  \loss_{\mathrm{geo}}
  =
  -\sum_i\sum_{j\neq i}
  w_{ij}
  \log
  \frac{\exp(\gtok_i^\top\gtok_j/\tau)}
       {\sum_{k\neq i}\exp(\gtok_i^\top\gtok_k/\tau)},
  \label{eq:geoce}
\end{equation}
with temperature $\tau$.
Since the target is soft, multiple patches can share high weight when they observe the same surface, making cross-view merging explicit rather than leaving it to the rendering loss.

\paragraph{Training.}
The model is trained end-to-end with rendering, geometric, and camera supervision:
\begin{equation}
  \loss
  =
  \loss_{\mathrm{render}}
  + \lambda_{\mathrm{geo}}\,\loss_{\mathrm{geo}}
  + \lambda_{\mathrm{cone}}\,\loss_{\mathrm{cone}}
  + \lambda_{\mathrm{pose}}\,\loss_{\mathrm{pose}}.
  \label{eq:loss}
\end{equation}
The rendering loss combines $\ell_1$ and perceptual terms on novel target views, while $\loss_{\mathrm{pose}}$ supervises the predicted cameras.
Coefficients and details are in \cref{sec:supp-training}.

\subsection{TokenBA: Refining Cameras and Scene Tokens}
\label{sec:tokenba}
The feed-forward prediction carries residual camera and geometry error.
The patch--Gaussian associations from \cref{sec:tokens} persist at inference and provide 2D--3D correspondences for refining both cameras and scene tokens from the input views within seconds.

\paragraph{Camera update.}
With the geometry fixed, we update each camera pose $\pose_n$ using the Gaussian--patch correspondences.
For each Gaussian $j$ of token $k$ with a supporting patch $i\in C_k$ visible in view $n$, we minimize the robust reprojection error:
\begin{equation}
  E_{\mathrm{cam}}(\pose_n)
  =
  \sum_{(k,j,i)}
  \rho\!\left(
  \left\|
  \pi(\pose_n,\gmean_{kj})-\*{u}_i
  \right\|
  \right),
  \label{eq:pnp}
\end{equation}
where $\rho$ is the penalty from \cref{eq:cone}.
Each pose is estimated independently by robust PnP, minimizing \cref{eq:pnp} over all its correspondences with graduated non-convexity, followed by a short photometric polish to reduce patch-center quantization.

\paragraph{Scene token update.}
With the cameras fixed, we refine the scene tokens by minimizing the rendering loss on the input views.
Gradients propagate through the renderer and frozen decoder to the tokens, combining evidence from all views.
Optimizing tokens instead of raw Gaussians retains the decoder as a learned scene prior.

\paragraph{Alternating optimization.}
Camera correction improves the consistency of rendering gradients, while token refinement improves the geometry for subsequent camera updates.
TokenBA alternates these steps and concludes with a short joint photometric refinement of tokens and poses, improving the feed-forward reconstruction within a few seconds (\cref{fig:losses-tokenba}).

\begin{figure*}[t]
\centering
\newlength{\qlw}\setlength{\qlw}{1.1em}%
\newlength{\qiw}\setlength{\qiw}{\dimexpr(\linewidth - \qlw)/6 - 2pt\relax}%
\newcommand{\qimg}[1]{\includegraphics[width=\qiw]{figures/qualitative/assets/#1}}%
\hspace{\qlw}%
\makebox[\qiw]{\small GT}\hfill%
\makebox[\qiw]{\small \ours}\hfill%
\makebox[\qiw]{\small \ours{} w/o tBA}\hfill%
\makebox[\qiw]{\small TokenGS}\hfill%
\makebox[\qiw]{\small YoNoSplat}\hfill%
\makebox[\qiw]{\small DA3}%
\\[0pt]
\begin{minipage}[c]{\qlw}%
\rotatebox[origin=c]{90}{\small 16 views}%
\end{minipage}%
\begin{minipage}[c]{\dimexpr\linewidth-\qlw\relax}%
\qimg{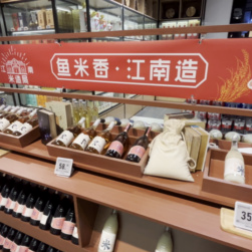}\hfill%
\qimg{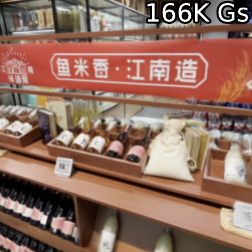}\hfill%
\qimg{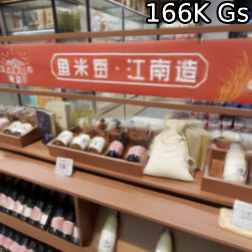}\hfill%
\qimg{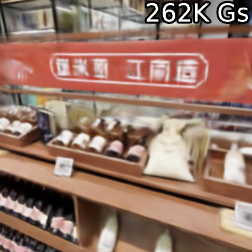}\hfill%
\qimg{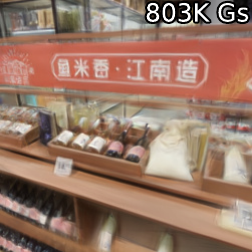}\hfill%
\qimg{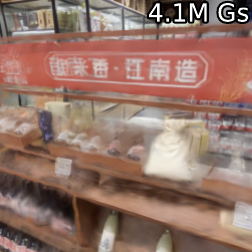}%
\end{minipage}%
\\[2pt]
\begin{minipage}[c]{\qlw}%
\rotatebox[origin=c]{90}{\small 32 views}%
\end{minipage}%
\begin{minipage}[c]{\dimexpr\linewidth-\qlw\relax}%
\qimg{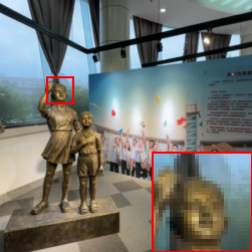}\hfill%
\qimg{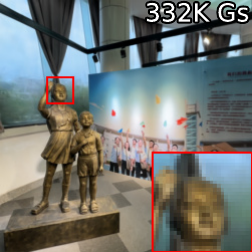}\hfill%
\qimg{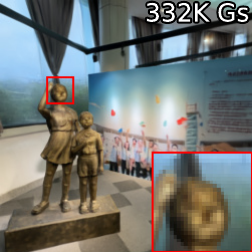}\hfill%
\qimg{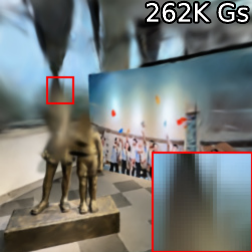}\hfill%
\qimg{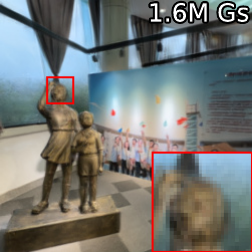}\hfill%
\qimg{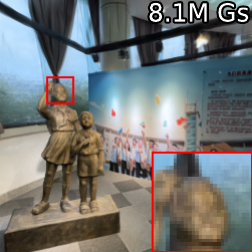}%
\end{minipage}%
\\[2pt]
\begin{minipage}[c]{\qlw}%
\rotatebox[origin=c]{90}{\small 64 views}%
\end{minipage}%
\begin{minipage}[c]{\dimexpr\linewidth-\qlw\relax}%
\qimg{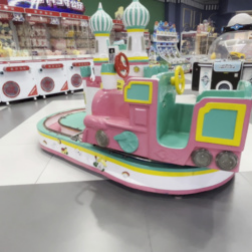}\hfill%
\qimg{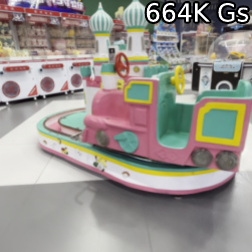}\hfill%
\qimg{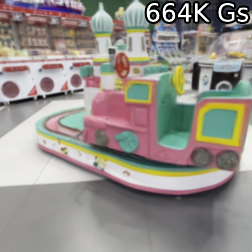}\hfill%
\qimg{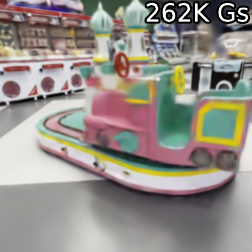}\hfill%
\qimg{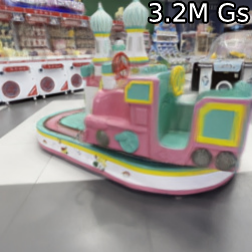}\hfill%
\qimg{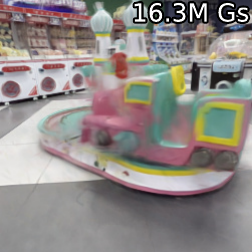}%
\end{minipage}%
\\[2pt]
\begin{minipage}[c]{\qlw}%
\rotatebox[origin=c]{90}{\small 128 views}%
\end{minipage}%
\begin{minipage}[c]{\dimexpr\linewidth-\qlw\relax}%
\qimg{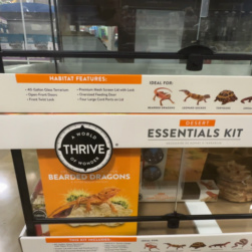}\hfill%
\qimg{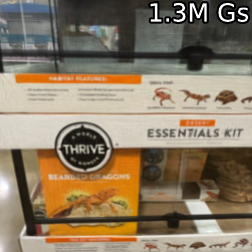}\hfill%
\qimg{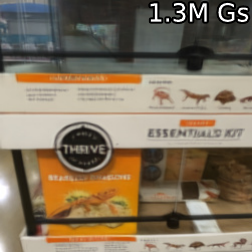}\hfill%
\qimg{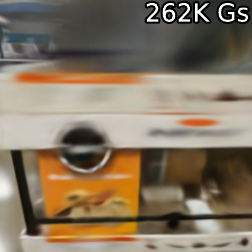}\hfill%
\qimg{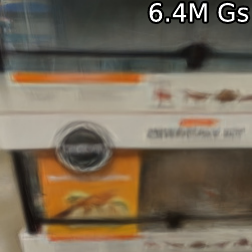}\hfill%
\qimg{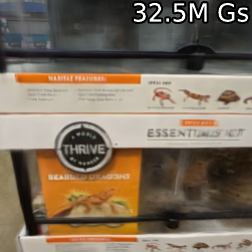}%
\end{minipage}%
\caption{
\textbf{Qualitative comparison on DL3DV} from 16 to 128 input views.
\ours{} maintains sharp, detailed renders as input coverage increases, outperforming both the token-based TokenGS and the pixel-aligned YoNoSplat and DA3.
TokenBA (tBA) refines the reconstruction, most visibly on fine detail like text and textures.
}
\label{fig:qualitative}
\end{figure*}

\section{Experiments}
\label{sec:experiments}

\paragraph{Implementation Details.}
We initialize the backbone and camera head from DA3-Giant~\citep{lin2025depth}.
Features from three backbone layers are merged into geometry tokens and fused into \stoknames{} through three cross- and self-attention blocks.
Each \stokname{} produces $\ngs{=}32$ Gaussians with degree-1 spherical harmonics.
We train on an equal mixture of RealEstate10K~\citep{zhou2018re10k} and DL3DV~\citep{ling2024dl3dv} at $252{\times}252$ resolution for 450K steps on 16 GH200 GPUs.
The input-view count increases from 2 to 24 over the first half of training.
Rendering supervision combines $\ell_1$ and LPIPS~\citep{zhang2018lpips} (weight $0.1$) on four target views, with auxiliary weights $\lambda_\text{geo}{=}0.05$, $\lambda_\text{cone}{=}0.01$, and $\lambda_\text{pose}{=}0.05$.
We sample the compression ratio uniformly from $[0.2/\nviews,1]$, giving a minimum budget of approximately 2K Gaussians.
TokenBA uses up to four alternating rounds followed by at most 96 joint refinement steps with an $\ell_1$ objective, adding approximately 1.6\,s per scene at 16 views on an RTX 4090.
Further details are provided in \cref{sec:supp-implementation}.

\subsection{Multi-View Novel View Synthesis}
\label{sec:exp-multiview}

\begin{table*}[t]
\centering
\caption{\textbf{Novel view synthesis} on DL3DV (16 and 64 views) and RealEstate10K (6 views).
Posed methods receive ground-truth cameras and intrinsics, others estimate them from images.
\colorbox{tabfirst}{Best}, \colorbox{tabsecond}{second}, \colorbox{tabthird}{third}.}
\label{tbl:nvs-main}
\scriptsize
\renewcommand{\arraystretch}{1.05}
\setlength\tabcolsep{2.5pt}
\resizebox{\linewidth}{!}{%
\begin{tabular}{lccccccccccccc}
\toprule
\multirow{2}{*}[-0.4em]{Method} &
& \multicolumn{4}{c}{DL3DV 16v}
& \multicolumn{4}{c}{DL3DV 64v}
& \multicolumn{4}{c}{RE10K 6v} \\
\cmidrule(lr){3-6} \cmidrule(lr){7-10} \cmidrule(lr){11-14}
& Posed & \#Gs & PSNR$\uparrow$ & SSIM$\uparrow$ & LPIPS$\downarrow$ & \#Gs & PSNR$\uparrow$ & SSIM$\uparrow$ & LPIPS$\downarrow$ & \#Gs & PSNR$\uparrow$ & SSIM$\uparrow$ & LPIPS$\downarrow$ \\
\midrule
DepthSplat~\citep{xu2025depthsplat} & \checkmark & 1.8M & 19.64 & \cthird 0.660 & \cthird 0.278 & 7.3M & 17.32 & 0.555 & 0.390 & 688K & 24.62 & \cthird 0.873 & \csecond 0.138 \\
ZPressor~\citep{wang2026zpressor} & \checkmark & 688K & 18.44 & 0.575 & 0.339 & 688K & 18.36 & 0.568 & 0.345 & 688K & 24.89 & 0.864 & \cthird 0.141 \\
VolSplat~\citep{wang2026volsplat} & \checkmark & 1.0M & 15.42 & 0.385 & 0.565 & 4.2M & 15.55 & 0.398 & 0.566 & 393K & 22.98 & 0.807 & 0.230 \\
GlobalSplat~\citep{itkin2026globalsplat} & \checkmark & 16K & 15.48 & 0.397 & 0.630 & 16K & 15.72 & 0.402 & 0.626 & 16K & 23.68 & 0.782 & 0.232 \\
TokenGS~\citep{ren2026tokengs} & \checkmark & 262K & \csecond 21.41 & \csecond 0.695 & 0.345 & 262K & \csecond 22.66 & \csecond 0.740 & 0.328 & 262K & \csecond 27.59 & \cfirst 0.899 & 0.145 \\
\arrayrulecolor{black!30}\cmidrule{1-14}\arrayrulecolor{black}
AnySplat~\citep{jiang2025anysplat} &  & 2.3M & 18.40 & 0.559 & 0.314 & 7.2M & 19.91 & 0.617 & 0.278 & 775K & 22.60 & 0.791 & 0.189 \\
EcoSplat~\citep{park2025ecosplat} &  & 419K & 14.05 & 0.319 & 0.605 & 1.7M & 13.06 & 0.275 & 0.647 & 157K & 21.73 & 0.728 & 0.267 \\
C3G~\citep{an2025c3g} &  & 2K & 12.71 & 0.286 & 0.649 & 2K & 11.56 & 0.243 & 0.672 & 2K & 20.64 & 0.663 & 0.318 \\
DA3~\citep{lin2025depth} &  & 4.1M & 20.34 & 0.652 & \csecond 0.276 & 16.3M & 20.59 & 0.666 & \csecond 0.257 & 1.5M & 21.37 & 0.726 & 0.229 \\
YoNoSplat~\citep{ye2026yonosplat} &  & 803K & \cthird 20.36 & 0.638 & 0.299 & 3.2M & \cthird 21.00 & \cthird 0.674 & \cthird 0.275 & 301K & \cthird 25.05 & 0.830 & 0.154 \\
\textbf{\ours} &  & 166K & \cfirst 24.79 & \cfirst 0.808 & \cfirst 0.175 & 664K & \cfirst 28.99 & \cfirst 0.900 & \cfirst 0.106 & 62K & \cfirst 28.50 & \csecond 0.882 & \cfirst 0.120 \\
\bottomrule
\end{tabular}}%
\end{table*}

\begin{table*}[t]
\centering
\vspace{-1mm}
\caption{\textbf{Cross-dataset generalization} to Mip-NeRF360 and ScanNet++.
Posed methods receive ground-truth cameras and intrinsics, others estimate them from images.
\colorbox{tabfirst}{Best}, \colorbox{tabsecond}{second}, \colorbox{tabthird}{third}.
*: trained on ScanNet++.}
\label{tbl:xd-combined}
\scriptsize
\renewcommand{\arraystretch}{1.05}
\setlength\tabcolsep{2.5pt}
\resizebox{\linewidth}{!}{%
\begin{tabular}{clccccccccccccc}
\toprule
& \multirow{2}{*}[-0.4em]{Method} &
& \multicolumn{4}{c}{32v}
& \multicolumn{4}{c}{64v}
& \multicolumn{4}{c}{128v} \\
\cmidrule(lr){4-7} \cmidrule(lr){8-11} \cmidrule(lr){12-15}
&& Posed & \#Gs & PSNR$\uparrow$ & SSIM$\uparrow$ & LPIPS$\downarrow$ & \#Gs & PSNR$\uparrow$ & SSIM$\uparrow$ & LPIPS$\downarrow$ & \#Gs & PSNR$\uparrow$ & SSIM$\uparrow$ & LPIPS$\downarrow$ \\
\midrule
\multirow{6}{*}[0.0em]{\rotatebox{90}{\tiny\textbf{Mip-NeRF}}}
& \color{black!50}3DGS$_{\text{30k}}$~\citep{kerbl20233d} & \color{black!50}\checkmark & \color{black!50}1.0M & \color{black!50}21.28 & \color{black!50}0.678 & \color{black!50}0.282 & \color{black!50}1.2M & \color{black!50}24.84 & \color{black!50}0.806 & \color{black!50}0.188 & \color{black!50}1.3M & \color{black!50}27.21 & \color{black!50}0.862 & \color{black!50}0.137 \\
\arrayrulecolor{black!30}\cmidrule{2-15}\arrayrulecolor{black}
& AnySplat~\citep{jiang2025anysplat} &  & 4.5M & 19.10 & 0.537 & \cthird 0.322 & 8.2M & 19.91 & 0.551 & \cthird 0.302 & 14.5M & 20.03 & 0.556 & \cthird 0.302 \\
& DA3~\citep{lin2025depth} &  & 8.1M & \cthird 21.48 & \cthird 0.603 & \csecond 0.266 & 16.3M & \cthird 21.03 & \cthird 0.585 & \csecond 0.277 & 32.5M & \cthird 20.62 & \cthird 0.579 & \csecond 0.287 \\
& TokenGS~\citep{ren2026tokengs} & \checkmark & 262K & \csecond 22.84 & \csecond 0.648 & 0.434 & 262K & \csecond 23.07 & \csecond 0.656 & 0.432 & 262K & \csecond 23.22 & \csecond 0.663 & 0.424 \\
& YoNoSplat~\citep{ye2026yonosplat} &  & 1.6M & 17.61 & 0.411 & 0.484 & 3.2M & 17.73 & 0.417 & 0.479 & 6.4M & 17.54 & 0.416 & 0.494 \\
& \textbf{\ours} &  & 332K & \cfirst 24.43 & \cfirst 0.696 & \cfirst 0.210 & 664K & \cfirst 25.24 & \cfirst 0.723 & \cfirst 0.185 & 1.3M & \cfirst 25.81 & \cfirst 0.739 & \cfirst 0.169 \\
\midrule
\multirow{6}{*}[0.0em]{\rotatebox{90}{\tiny\textbf{ScanNet++}}}
& \color{black!50}3DGS$_{\text{30k}}$~\citep{kerbl20233d} & \color{black!50}\checkmark & \color{black!50}1.2M & \color{black!50}20.11 & \color{black!50}0.690 & \color{black!50}0.365 & \color{black!50}1.6M & \color{black!50}21.16 & \color{black!50}0.732 & \color{black!50}0.345 & \color{black!50}1.7M & \color{black!50}22.14 & \color{black!50}0.775 & \color{black!50}0.324 \\
\arrayrulecolor{black!30}\cmidrule{2-15}\arrayrulecolor{black}
& AnySplat*~\citep{jiang2025anysplat} &  & 3.5M & 16.63 & 0.669 & \csecond 0.391 & 6.9M & 18.36 & 0.694 & \csecond 0.348 & 12.8M & 18.53 & 0.692 & \csecond 0.354 \\
& DA3*~\citep{lin2025depth} &  & 8.1M & \cthird 18.48 & \cthird 0.683 & \cthird 0.394 & 16.3M & \cthird 19.15 & \cthird 0.702 & \cthird 0.362 & 32.5M & \cthird 19.12 & \cthird 0.703 & \cthird 0.361 \\
& TokenGS~\citep{ren2026tokengs} & \checkmark & 262K & \csecond 22.14 & \csecond 0.772 & 0.422 & 262K & \csecond 22.49 & \csecond 0.782 & 0.426 & 262K & \csecond 22.71 & \csecond 0.786 & 0.427 \\
& YoNoSplat~\citep{ye2026yonosplat} &  & 1.6M & 16.13 & 0.622 & 0.516 & 3.2M & 16.29 & 0.626 & 0.506 & 6.4M & 16.12 & 0.623 & 0.513 \\
& \textbf{\ours} &  & 332K & \cfirst 22.36 & \cfirst 0.779 & \cfirst 0.298 & 664K & \cfirst 24.36 & \cfirst 0.826 & \cfirst 0.249 & 1.3M & \cfirst 25.40 & \cfirst 0.848 & \cfirst 0.235 \\
\bottomrule
\end{tabular}}%
\end{table*}

\paragraph{Setup.}
We evaluate on DL3DV~\citep{ling2024dl3dv} with 16 and 64 input views, and on RealEstate10K~\citep{zhou2018re10k} with 6 input views.
Target views are held out across the full sequence, and context views are chosen from the remaining frames to maximize coverage (\cref{sec:supp-eval-views}).
Each method runs inference at their training resolution, and all renderings are evaluated at $252{\times}252$ (\cref{sec:supp-eval-resolution}).
All results use full capacity ($\qr{=}1$).
Each method runs its full released inference setting: TokenGS applies test-time token optimization (TTO) and \ours{} applies TokenBA.

\paragraph{Baselines.}
We compare against pose-free methods AnySplat~\citep{jiang2025anysplat}, EcoSplat~\citep{park2025ecosplat}, DA3~\citep{lin2025depth}, C3G~\citep{an2025c3g}, and YoNoSplat~\citep{ye2026yonosplat}.
Baselines using ground-truth cameras include the pixel-aligned DepthSplat~\citep{xu2025depthsplat} and ZPressor~\citep{wang2026zpressor}, voxel-based VolSplat~\citep{wang2026volsplat}, and token-based GlobalSplat~\citep{itkin2026globalsplat} and TokenGS~\citep{ren2026tokengs}.
DA3 uses the same backbone as \ours{}, providing a comparison between pixel-aligned prediction and our scene-token formulation with a shared backbone.

\paragraph{Results.}
\Cref{tbl:nvs-main} shows that \ours{} achieves the highest quality in all three settings, including against methods given ground-truth cameras.
On DL3DV, the advantage grows with more views: from $4.45$\,dB over DA3, which shares our backbone, at 16 views to $8.40$\,dB at 64 views, with about $24{\times}$ fewer Gaussians.
Pixel-aligned prediction ties each Gaussian to a single view's depth and pose errors; as views increase, these misalignments accumulate rather than cancel, whereas our free 3D placement avoids this dependence (\cref{fig:qualitative}).
Against TokenGS with TTO, the strongest token-based baseline, the gap widens from $3.38$ to $6.33$\,dB across the same range.
Despite its strong PSNR, TokenGS blurs fine details and misses scene content (\cref{fig:qualitative,fig:re10k-qualitative}), because learned queries must infer their observation support implicitly.
Our scene tokens start from their supporting patches, preserving content and producing sharper reconstructions.
On RealEstate10K, \ours{} exceeds TokenGS by $0.91$\,dB with $4{\times}$ fewer Gaussians; posed methods lead only in SSIM, likely benefiting from ground-truth intrinsics.
Even without TokenBA, \ours{} outperforms all evaluated feed-forward baselines in PSNR and LPIPS on every benchmark (\cref{sec:supp-full-dl3dv}).

\subsection{Cross-Dataset Generalization}
\label{sec:exp-crossdataset}

\begin{figure*}[t]
\centering
\ifdefined\xdlw\else\newlength{\xdlw}\fi\setlength{\xdlw}{1.1em}%
\ifdefined\xdin\else\newlength{\xdin}\fi\setlength{\xdin}{2pt}%
\ifdefined\xdgap\else\newlength{\xdgap}\fi\setlength{\xdgap}{6pt}%
\ifdefined\xdiw\else\newlength{\xdiw}\fi\setlength{\xdiw}{\dimexpr(\linewidth - \xdlw - 3\xdin - 2\xdgap - 1pt)/6\relax}%
\newcommand{\xdimg}[1]{\includegraphics[width=\xdiw]{figures/xd_qualitative/assets/#1}}%
\hspace{\xdlw}%
\makebox[\dimexpr2\xdiw+\xdin\relax]{\small 32 views}\hspace{\xdgap}%
\makebox[\dimexpr2\xdiw+\xdin\relax]{\small 64 views}\hspace{\xdgap}%
\makebox[\dimexpr2\xdiw+\xdin\relax]{\small 128 views}%
\\[0pt]
\begin{minipage}[c]{\xdlw}%
\rotatebox[origin=c]{90}{\small Mip-NeRF}%
\end{minipage}%
\begin{minipage}[c]{\dimexpr\linewidth-\xdlw\relax}%
\xdimg{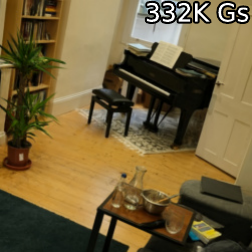}\hspace{\xdin}%
\xdimg{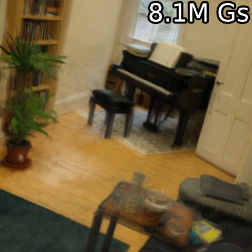}\hspace{\xdgap}%
\xdimg{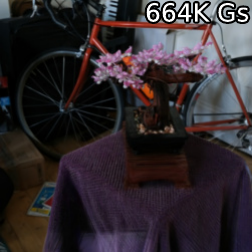}\hspace{\xdin}%
\xdimg{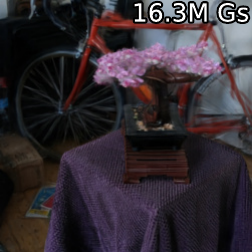}\hspace{\xdgap}%
\xdimg{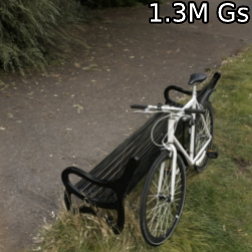}\hspace{\xdin}%
\xdimg{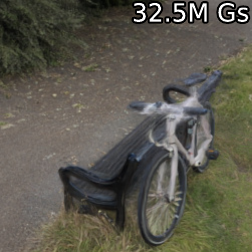}%
\end{minipage}%
\\[1pt]
\begin{minipage}[c]{\xdlw}%
\rotatebox[origin=c]{90}{\small ScanNet++}%
\end{minipage}%
\begin{minipage}[c]{\dimexpr\linewidth-\xdlw\relax}%
\xdimg{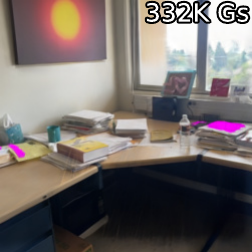}\hspace{\xdin}%
\xdimg{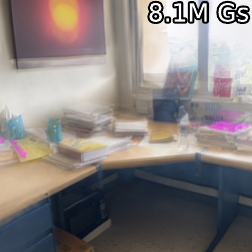}\hspace{\xdgap}%
\xdimg{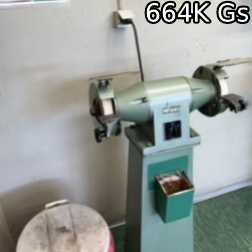}\hspace{\xdin}%
\xdimg{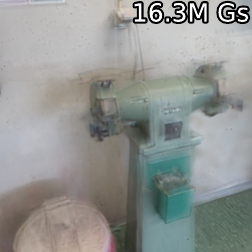}\hspace{\xdgap}%
\xdimg{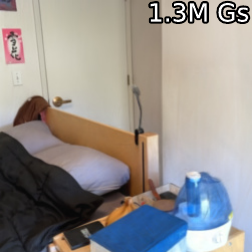}\hspace{\xdin}%
\xdimg{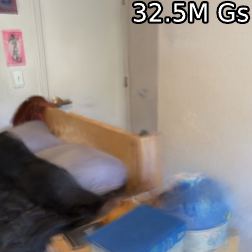}%
\end{minipage}%
\\[1pt]
\hspace{\xdlw}%
\makebox[\xdiw]{\small \ours}\hspace{\xdin}%
\makebox[\xdiw]{\small DA3}\hspace{\xdgap}%
\makebox[\xdiw]{\small \ours}\hspace{\xdin}%
\makebox[\xdiw]{\small DA3}\hspace{\xdgap}%
\makebox[\xdiw]{\small \ours}\hspace{\xdin}%
\makebox[\xdiw]{\small DA3}%
\caption{\textbf{Cross-dataset qualitative comparison} on Mip-NeRF360 (top) and ScanNet++ (bottom) at 32, 64, and 128 input views.
For each view count we show \ours{} (left) and DA3 (right).
\ours{} produces sharper geometry and fewer artifacts, with quality improving steadily as more views become available.
}
\label{fig:xd-qualitative}
\end{figure*}

We evaluate zero-shot generalization on Mip-NeRF360~\citep{barron2022mipnerf360} and ScanNet++~\citep{yeshwanth2023scannetpp}.
Mip-NeRF360 contains indoor and outdoor scenes, while ScanNet++ covers cluttered indoor rooms.
We follow the sampling protocol used in \cref{sec:exp-multiview}.

We compare against the main pose-free methods, AnySplat, DA3, and YoNoSplat, and against TokenGS as the token-based reference.
AnySplat and DA3 train on ScanNet++, so their results there are not zero-shot.
Per-scene 3DGS~\citep{kerbl20233d} with ground-truth cameras and 30K iterations provides an optimization-based reference.

\Cref{tbl:xd-combined} and \cref{fig:xd-qualitative} show that \ours{} outperforms every pose-free baseline and TokenGS, which requires ground-truth cameras, across all metrics and view counts on both datasets.
Its quality rises with denser input, by $1.38$\,dB on Mip-NeRF360 and $3.04$\,dB on ScanNet++ from 32 to 128 views, while the baselines largely plateau or decline.
Even without cameras, \ours{} exceeds per-scene 3DGS on ScanNet++ by $3.26$\,dB and comes within $1.40$\,dB of it on Mip-NeRF360 at 128 views.

\subsection{Adjustable Gaussian Budget}
\label{sec:exp-compression}

\begin{figure*}[t]
\centering
\begin{minipage}[t]{0.52\linewidth}
\vspace{0pt}%
\centering
\includegraphics[width=\linewidth]{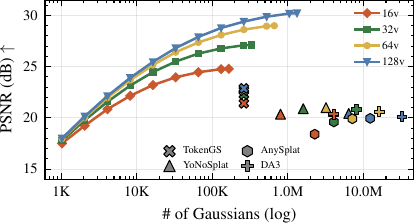}
\end{minipage}%
\hfill
\begin{minipage}[t]{0.46\linewidth}
\vspace{0pt}%
\centering
\ifdefined\ciw\else\newlength{\ciw}\fi%
\setlength{\ciw}{\dimexpr\linewidth/4 - 2pt\relax}%
\newcommand{\cimg}[1]{\includegraphics[width=\ciw]{figures/re10k_qualitative/assets/#1}}%
\cimg{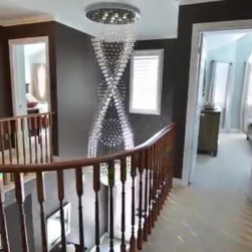}\hfill\cimg{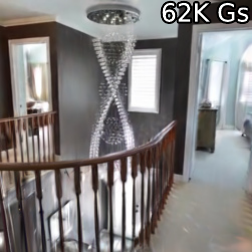}\hfill\cimg{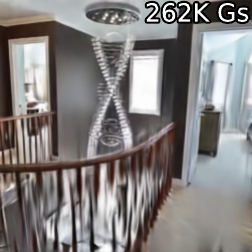}\hfill\cimg{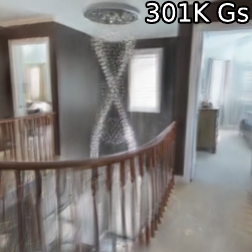}\\[1pt]%
\cimg{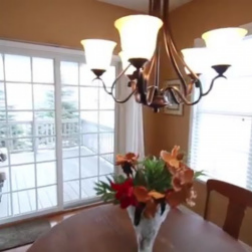}\hfill\cimg{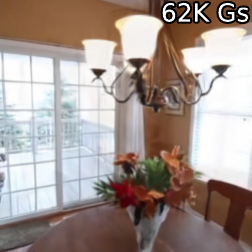}\hfill\cimg{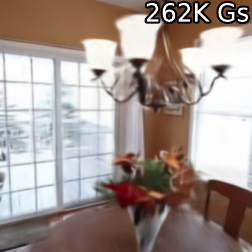}\hfill\cimg{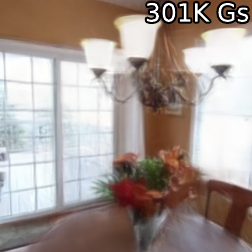}\\[1pt]%
\makebox[\ciw]{\small GT}\hfill%
\makebox[\ciw]{\small \ours}\hfill%
\makebox[\ciw]{\small TokenGS}\hfill%
\makebox[\ciw]{\small YoNoSplat}
\end{minipage}
\\[0pt]
\begin{minipage}[t]{0.52\linewidth}
\captionof{figure}{\textbf{Quality \vs Gaussian budget} on DL3DV.
Each \ours{} curve varies the compression ratio $\qr$ at inference.
More input views raise the curve at every budget.
}
\label{fig:compression-curve}
\end{minipage}%
\hfill
\begin{minipage}[t]{0.46\linewidth}
\captionof{figure}{\textbf{Comparison on RealEstate10K} (6 views).
\ours{} reconstructs fine details more faithfully than the baselines with fewer Gaussians.
}
\label{fig:re10k-qualitative}
\end{minipage}
\end{figure*}

Since the compression ratio $\qr$ is an inference-time parameter, one trained model produces reconstructions at any point of the quality--budget curve.
\Cref{fig:compression-curve} traces this curve on DL3DV for 16 to 128 input views, from 1K Gaussians up to full capacity.
Quality degrades gracefully: at 16 views, reducing the budget from 166K to 65K costs only $0.34$\,dB with about $2.5{\times}$ fewer Gaussians.
Compressing further to 8K Gaussians still exceeds every baseline, with $100{\times}$ fewer Gaussians than YoNoSplat and $500{\times}$ fewer than DA3 at 16 views.
The smallest budget of 1K Gaussians lies below the smallest training budget of about 65 \stoknames{}, yet the curve continues smoothly.
Additional views raise the entire curve rather than only its endpoint.
At a fixed budget of 8K Gaussians, PSNR rises from 16 to 128 views by $1.76$\,dB, so quality follows the observations rather than the number of Gaussians.
In \cref{fig:compression}, \stoknames{} cover larger regions as $\qr$ decreases while the render stays sharp.

\subsection{Pose Estimation and TokenBA}
\label{sec:exp-tokenba}

\begin{figure*}[t]
\centering
\includegraphics[width=\linewidth]{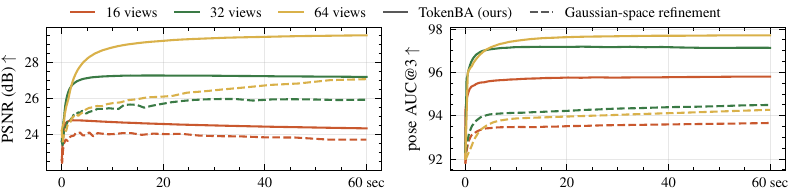}
\caption{\textbf{TokenBA \vs Gaussian-space refinement} on DL3DV.
Target-view PSNR (left) and pose accuracy (right) vs. refinement time.
TokenBA converges faster and to a higher level than direct Gaussian optimization.
}
\label{fig:tokenba}
\end{figure*}

\providecommand{\posefrac}{0.42}
\begin{table*}[t]
\begin{minipage}[t]{\posefrac\linewidth}
\vspace{0pt}
\centering
\captionof{table}{\textbf{Pose estimation}.
AUC of relative pose error.
We color the \colorbox{tabfirst}{best} and \colorbox{tabsecond}{second best}.}
\label{tbl:pose}
\scriptsize
\renewcommand{\arraystretch}{1.05}
\setlength\tabcolsep{2.5pt}
\resizebox{\linewidth}{!}{%
\begin{tabular}{lcccccc}
\toprule
Method
& \multicolumn{3}{c}{DL3DV 64v}
& \multicolumn{3}{c}{RE10K 6v} \\
\cmidrule(lr){2-4} \cmidrule(lr){5-7}
\multicolumn{1}{r}{AUC$\uparrow$} & $1^\circ$ & $3^\circ$ & $5^\circ$ & $1^\circ$ & $3^\circ$ & $5^\circ$ \\
\midrule
AnySplat~\citep{jiang2025anysplat} & 0.071 & 0.357 & 0.550 & 0.130 & 0.378 & 0.530 \\
DA3~\citep{lin2025depth} & \csecond 0.930 & \csecond 0.969 & \csecond 0.980 & \csecond 0.546 & \csecond 0.741 & \csecond 0.817 \\
YoNoSplat~\citep{ye2026yonosplat} & 0.466 & 0.754 & 0.842 & 0.361 & 0.673 & 0.779 \\
\textbf{\ours} & \cfirst 0.950 & \cfirst 0.976 & \cfirst 0.983 & \cfirst 0.738 & \cfirst 0.855 & \cfirst 0.897 \\
\bottomrule
\end{tabular}}%

\end{minipage}%
\hfill
\begin{minipage}[t]{\dimexpr0.98\linewidth-\posefrac\linewidth\relax}
\vspace{0pt}
\centering
\captionof{table}{\textbf{Ablations} on DL3DV (6 views).
From per-pixel to fully grounded \stoknames{}.}
\label{tbl:ablations}
\scriptsize
\renewcommand{\arraystretch}{1.05}
\setlength\tabcolsep{2.5pt}
\resizebox{\linewidth}{!}{%
\begin{tabular}{lccccccccc}
\toprule
\multirow{2}{*}[-0.4em]{Method} & & & &
& \multicolumn{2}{c}{$\qr{=}1$}
& \multicolumn{2}{c}{$0.5$}
& \multicolumn{1}{c}{AUC@3$\uparrow$} \\
\cmidrule(lr){6-7} \cmidrule(lr){8-9} \cmidrule(lr){10-10}
& Adj.\ budget & Cone & GeoCE & TokenBA & \#Gs & PSNR$\uparrow$ & \#Gs & PSNR$\uparrow$ & $\qr{=}1$ \\
\midrule
Per-pixel & -- & -- & -- & -- & 381K & 22.62 & -- & -- & 0.542 \\
Fixed queries & {\footnotesize$\times$} & frustum & -- & -- & 262K & 23.02 & -- & -- & 0.632 \\
\arrayrulecolor{black!30}\cmidrule{1-10}\arrayrulecolor{black}
Obs. tokens & \checkmark & frustum & -- & -- & 62K & 22.87 & 31K & 22.53 & 0.615 \\
+ Cone & \checkmark & \checkmark & -- & -- & 62K & 24.38 & 31K & 23.09 & 0.669 \\
+ GeoCE & \checkmark & \checkmark & \checkmark & -- & 62K & 24.34 & 31K & 23.25 & 0.686 \\
+ TokenBA & \checkmark & \checkmark & \checkmark & \checkmark & 62K & 27.79 & 31K & 27.13 & 0.889 \\
\bottomrule
\end{tabular}}%

\end{minipage}
\end{table*}

\Cref{tbl:pose} reports pose accuracy as the AUC of the relative pose error at $1^\circ$, $3^\circ$, and $5^\circ$.
\ours{} achieves the highest accuracy at every threshold on both datasets.
Compared with DA3, AUC at $1^\circ$ increases from $0.930$ to $0.950$ on DL3DV and from $0.546$ to $0.738$ on RealEstate10K.

\Cref{fig:tokenba} compares TokenBA against Gaussian-space refinement~\citep{fan2024instantsplat} on DL3DV, running both for $60$\,s without early stopping.
TokenBA converges faster and reaches higher quality within this budget: at 16 views, $90\%$ of its gain arrives within $0.4$\,s, and its margin over Gaussian-space refinement grows from $0.6$\,dB at 16 views to $2.4$\,dB at 64.
The first camera update alone lifts AUC at $3^\circ$ from $91.8$ to $94.3$ within milliseconds, beyond what Gaussian-space refinement reaches in $60$\,s ($93.7$).
All other results use the early-stopping schedule, which stops after $1.6$ to $11.7$\,s.
The token--patch correspondences give TokenBA explicit 2D--3D associations that photometric-only refinement lacks, improving not only convergence speed but also accuracy and robustness.
\subsection{Ablations}
\label{sec:exp-ablations} 

\Cref{tbl:ablations} ablates design choices on DL3DV at 6 views; every row is a DA3-Base retrain with the same training budget.
Per-pixel prediction gives $22.62$\,dB from $381$K Gaussians with no budget control.
Fixed learned queries (matching TokenGS~\citep{ren2026tokengs}: $4$K queries $\times$ $64$ Gaussians) reach $23.02$\,dB at $262$K Gaussians with their frustum loss, which constrains Gaussians to lie within the input views, but the budget is locked at training time.
Observation-derived tokens make the budget adjustable at inference: $62$K Gaussians reach $22.87$\,dB, and halving to $31$K costs only $0.34$\,dB.
Yet the frustum loss does not tie tokens to their source observations.
The cone loss grounds them and adds the largest training-time gain ($+1.51$\,dB).
It also provides the 2D--3D correspondences TokenBA relies on: without it, TokenBA collapses to $14.7$\,dB.
GeoCE further improves clustering by making tokens that observe the same surface merge, adding $+0.16$\,dB at $\qr{=}0.5$ and lifting AUC@3 from $0.669$ to $0.686$.
With grounding established, TokenBA refines cameras and tokens through the learned correspondences, delivering the large gains in both reconstruction and pose accuracy.

\section{Conclusion}
\label{sec:conclusion}

We present \ours{}, a feed-forward model that reconstructs scenes from unposed images as compact, observation-grounded \stoknames{}.
Each token is derived from an image patch and decoded into freely placed 3D Gaussians.
Tokens across views are merged, removing redundancy and making the budget adjustable.
The cone loss ties each token's Gaussians to its source patch, which is what makes the grounding pay off.
GeoCE shapes the token space so that merged tokens observe the same surface.
The resulting correspondences enable TokenBA, which refines cameras and tokens within seconds.
Without poses or intrinsics, \ours{} outperforms baselines with substantially fewer Gaussians, including those given ground-truth cameras, and generalizes zero-shot to unseen datasets.

\paragraph{Limitations:}
The camera estimates can degrade on scenes that differ strongly from the training distribution.
Ground-truth cameras close this gap, and TokenBA recovers most of it, but a residual remains on out-of-distribution captures.
The method also assumes static scenes.

\ificlrfinal
\subsubsection*{Acknowledgements}
We sincerely thank Shaohui Liu, Yung-Hsu Yang, and Haofei Xu for their helpful feedback and discussions.
This work was supported under project ID a144 as part of the Swiss AI Initiative, through a grant from the ETH Domain and computational resources provided by the Swiss National Supercomputing Centre (CSCS) under the Alps infrastructure.
\fi

\subsection*{AI use statement}

We used generative AI tools to assist with implementing methods and experiments, interpreting results, and drafting and editing the paper.
All ideas behind the method are our own, and we have not used generative AI tools to generate data.
We have reviewed and verified all AI-assisted work, and all reported numbers were reproduced with our own evaluation runs.
We take responsibility for the final content of this work, including text, claims or artifacts produced with the aid of generative AI.

\subsection*{Reproducibility statement}

\Cref{sec:experiments} and \cref{sec:supp-implementation} give the implementation details of \ours{}, including the architecture, the losses, the training schedule, and the TokenBA settings.
Training and evaluation use only public datasets, and \cref{sec:supp-evaluation} documents the evaluation protocol, including the view selection (\cref{sec:supp-eval-views}).
We re-evaluate all baselines under this same protocol, with the roster in \cref{sec:supp-eval-baselines} and the full results in \cref{sec:supp-full-results}.
\ificlrfinal The code and trained models are available on our project page.\else We will release the code and the trained models.\fi

\bibliographystyle{iclr2027_conference}
\bibliography{mybib}

\ifaddsupp
\clearpage
\appendix

\section{Additional Implementation Details}
\label{sec:supp-implementation}

\subsection{Architecture}
\label{sec:supp-architecture}

\paragraph{Backbone features.}
The backbone is a pretrained DA3-Giant~\citep{lin2025depth} (40 layers, width $1536$, 24 heads of dimension 64).
We extract features from layers 19, 29, and 39, each comprising a local component from within-view attention and a global component incorporating cross-view information.

\paragraph{Geometry tokens.}
Clustering requires a space where proximity reflects shared 3D structure.
The backbone produces per-patch features at multiple layers, each with a local component $\vtok_l^\text{loc}$ from within-view attention and a global component $\vtok_l^\text{glob}$ from cross-view attention.
These are fused per layer as
\begin{equation}
  \vtok_l = \vtok_l^\text{glob} + \alpha_l \cdot \operatorname{LN}(\vtok_l^\text{loc}),
  \label{eq:payload}
\end{equation}
where $\alpha_l$ is a learnable gate initialized to $0.1$.
A separate two-layer MLP embeds each raw $14{\times}14$ RGB patch into a $1536$-dimensional color token $\vtok^\text{col}$.
All streams merge into a single geometry token per patch,
\begin{equation}
  \gtok = \sum_{l} \beta_l \cdot \operatorname{MLP}_l\!\bigl(\operatorname{LN}(\vtok_l)\bigr) + \beta_c \cdot \operatorname{MLP}_c\!\bigl(\operatorname{LN}(\vtok^\text{col})\bigr),
  \label{eq:merge}
\end{equation}
with learnable scales $\beta_l, \beta_c$ initialized to $1/4$ and two-layer GELU MLPs.
This shared space defines the clustering metric, the cross-attention keys, and the \stokname{} initialization.
GeoCE supervises these tokens directly; the per-layer payloads and color embedding supply the attention values.

\paragraph{Fusion and decoding.}
Fusion uses three blocks.
Each block is a cross-attention (queries: \stoknames{}; keys: geometry tokens; values: the \vtoknames{} of one layer, deepest first) followed by self-attention among the \stoknames{}.
Blocks use 24 heads, MLP ratio 4, GELU, and residual gates initialized to $0.01$.
A final residual cross-attention with the color embedding as values adds appearance.
A two-layer MLP decodes each \stokname{} into $\ngs{=}32$ Gaussians of 23 parameters.

\paragraph{Reference frame.}
All geometry is expressed in the coordinate frame of the first input camera.
The scene scale is normalized so that the median distance of the input depth points to the origin is one.

\paragraph{Camera estimation and priors.}
A two-layer ReLU MLP maps the deepest backbone camera token to a translation, unit quaternion, and field of view, initialized from the DA3 camera head.
$\loss_{\mathrm{pose}}$ applies a Huber loss on translation, an $\ell_1$ chordal loss on rotations, and an $\ell_1$ loss on the field of view.
When ground-truth cameras are available, DA3's frozen camera encoder injects them into the camera token; this path is used with probability $0.2$ during training.

\paragraph{Gaussian parameterization.}
Positions use an inverse-log activation $\operatorname{sign}(y)(\exp|y|-1)$ on raw outputs clamped to $[-5,5]$, scales use $\softplus(x-4)$ clamped to $[10^{-6},15]$, rotations are normalized quaternions, opacities use a sigmoid with bias $-1.5$, and colors are degree-1 spherical harmonics (12 coefficients).
All 32 Gaussians of a token start from a shared template with a forward bias of $z{=}0.5$.
Squared-hinge penalties keep scales in $[10^{-6},0.1]$, opacities above $0.01$, and positions within a radius of 10.
\Cref{tbl:supp-gpp} varies the number of Gaussians per \stokname{}.
$\ngs{=}32$ gains $0.34$ to $0.36$\,dB over $\ngs{=}8$, while $\ngs{=}64$ adds at most $0.08$\,dB at twice the Gaussians, so \ours{} uses $\ngs{=}32$.
\begin{center}
\centering
\captionof{table}{\textbf{Gaussians per \stokname{}} on DL3DV at 6 views.
Each row is a DA3-Base model retrained to decode $\ngs$ Gaussians per \stokname{}, evaluated without TokenBA.
Columns compare full capacity ($\qr{=}1$) with compression ratio $\qr{=}0.5$, where half of the \stoknames{} are kept.
\#Gs is the total number of Gaussians.}
\label{tbl:supp-gpp}
\scriptsize
\renewcommand{\arraystretch}{1.05}
\setlength\tabcolsep{2.5pt}
\begin{tabular}{lcccc}
\toprule
\multirow{2}{*}[-0.4em]{$\ngs$}
& \multicolumn{2}{c}{$\qr{=}1$}
& \multicolumn{2}{c}{$\qr{=}0.5$} \\
\cmidrule(lr){2-3} \cmidrule(lr){4-5}
& \#Gs & PSNR$\uparrow$ & \#Gs & PSNR$\uparrow$ \\
\midrule
8 & 16K & 23.98 & 8K & 22.91 \\
16 & 31K & 24.20 & 16K & 23.11 \\
32 & 62K & 24.34 & 31K & 23.25 \\
64 & 124K & 24.42 & 62K & 23.33 \\
\bottomrule
\end{tabular}

\end{center}

\subsection{Geometric Losses}
\label{sec:supp-losses}

\paragraph{GeoCE loss.}
\label{sec:supp-geoce}
During training, 2048 tokens per scene are sampled as anchors, with each anchor's 3D point obtained by unprojecting its patch's median-depth pixel.
The soft target weight of \cref{eq:geoce_target} is averaged over a $4{\times}4$ pixel grid inside each key patch, and patches with valid depth on fewer than half their pixels are excluded.
The base bandwidth $\sigma$ is the scene-wise median 3D distance between adjacent patch points.
Per token, $\sigma_i$ is log-interpolated from $\sigma$ (flattest patch) to $0.4\,\sigma$ (most textured), ranked by mean image-gradient energy.
The temperature is $\tau{=}0.1$ and the loss acts on $\ell_2$-normalized geometry tokens with weight $\lambda_{\mathrm{geo}}{=}0.05$.
Anchor depths are precomputed by DA3-Giant from the ground-truth cameras of each training scene.

\paragraph{Cone loss.}
\label{sec:supp-cone}
The robust penalty $\rho$ in \cref{eq:cone} is a hinged Cauchy function of the reprojection distance in patch units.
The hinge sets a zero-cost region within 2 patches, allowing Gaussians to extend past their observations to cover flat surfaces or occluded regions.
Beyond the hinge, the penalty ramps smoothly ($C^1$) and saturates past 4 patches to limit the influence of distant outliers.
Gaussians that project behind a member view receive a sentinel distance of 20 patches with no lateral gradient.
The loss constrains only the lateral position relative to the supporting patches' viewing cones and leaves depth to the rendering loss, with weight $\lambda_{\mathrm{cone}}{=}0.01$.

\subsection{Clustering}
\label{sec:supp-clustering}

We run spherical $k$-means on the $\ell_2$-normalized geometry tokens with cosine assignments and centroid renormalization after each update.
Lloyd iterations are capped at 50 but typically converge in 2 to 3 iterations at $\qr{=}0.5$ and 7 to 9 at $\qr{=}0.1$.
Seeds are sampled without replacement with probability proportional to $(0.05 + c_i)^2$, where $c_i \in [0,1]$ is the per-scene rank of patch $i$ by mean image-gradient energy.
This biases initialization toward textured regions but does not change the objective.
Each \stokname{} is the mean of its members' geometry tokens, recomputed differentiably from the detached assignments.
During training, $\qr \sim \mathcal{U}[\qr_{\min}, 1]$ with $\qr_{\min}{=}0.2/\nviews$, giving at least about 65 \stoknames{} at any view count.
A fixed Gaussian budget $B$ corresponds to $\qr = B/(\ngs \nviews T)$.

\subsection{Training}
\label{sec:supp-training}

Training is a single stage from DA3-Giant on a 50/50 mixture of RealEstate10K~\citep{zhou2018re10k} and DL3DV~\citep{ling2024dl3dv} at $252{\times}252$.
It runs for 450K steps on 16 GH200 GPUs with 24 samples per GPU and 4 target views per sample.
We use AdamW ($\beta_1{=}0.9$, $\beta_2{=}0.95$, weight decay $0.05$, gradient clipping $1.0$) with a base learning rate of $3{\times}10^{-4}$, the backbone at $0.1{\times}$, $5\%$ linear warmup, cosine decay to zero, and bfloat16 autocast.
The input view count grows from 2 to 24 over the first half of training.
A single checkpoint, trained on the RealEstate10K and DL3DV mixture, is used across all benchmarks for evaluation.

\subsection{TokenBA}
\label{sec:supp-tokenba}

TokenBA optimizes the \stoknames{} $\stokset$ and the input camera poses under an $\ell_1$ rendering loss on the input views, with the decoder frozen.
Tokens use Adam with a learning rate of $10^{-3}$ during the alternation and $2{\times}10^{-3}$ in the joint tail.
The decoder MLP body runs in half precision while its readout stays in full precision.

\paragraph{Camera update.}
The correspondences in \cref{eq:pnp} follow from the cluster assignments: every Gaussian of a \stokname{} is paired with the patch center of each member patch.
Each pose is solved by robust PnP with graduated non-convexity, annealing the robust scale from 64 to 12 to 4 pixels over 4 iterations per level on 2048 sampled correspondences.
Seven rotation hypotheses per view (the current rotation and $\pm40^\circ$ about each camera axis) are solved in one batched call, and the hypothesis with the most inliers within 2 patches wins.
A photometric polish then removes the patch-center quantization error, running two steps at half resolution and one at full resolution with the aligner described below.

\paragraph{Photometric aligner.}
The aligner is a direct Levenberg--Marquardt solver on the photometric residual.
Each step renders at the current pose and takes a Gauss--Newton update on the brightness-constancy linearization with the analytic image-warp Jacobian, using one render per step and no autograd.
The full configuration uses a four-level pyramid ($1/8$, $1/4$, $1/2$, and full resolution) with 12, 10, 8, and 6 steps per level, Gaussian blur at the two coarse levels, and accept-reject steps with a trust-region clamp on the pose twist.
A scan over nine log-spaced global scale factors from $0.5$ to $2.0$ precedes the pyramid, and the best full-resolution pose across levels is kept.
Inside TokenBA only the two fine levels run, without verification or scale scan.

\paragraph{Token update and joint refinement.}
Each alternation round ends with one token refit step.
Rounds stop once the PnP snap moves every pose by less than $0.3^\circ$, or after four rounds.
The joint tail refines tokens and poses (an $\mathfrak{se}(3)$ increment with learning rate $3{\times}10^{-4}$) for at most 96 steps, reusing the same Adam state.
It exits once a block of 8 steps improves the loss by less than $2\%$.

\section{Evaluation Protocol}
\label{sec:supp-evaluation}

We describe the evaluation setup shared by all experiments: view selection, pose metrics, baselines, and the resolution and resize choices that affect reported numbers.

\subsection{View selection}
\label{sec:supp-eval-views}
On DL3DV (140 scenes), every 8th frame is held out as a target, and context views are selected from the remaining frames by $k$-means clustering on the 6-D feature vector of camera position and viewing direction~\citep{ziwen2025long}.
On Mip-NeRF360 (6 scenes) and ScanNet++ (46 iPhone captures), targets are selected by farthest point sampling over the camera centers, and context views over the remaining frames.
We evaluate at 16, 32, 64, and 128 input views on DL3DV and at 32, 64, and 128 on Mip-NeRF360 and ScanNet++, with targets fixed across view counts.
On RealEstate10K (1,600 scenes) we use the 6-view protocol of YoNoSplat~\citep{ye2026yonosplat} with 8 held-out targets.

\subsection{Pose accuracy}
\label{sec:supp-eval-pose}
We report relative rotation error and translation-direction error over all input-view pairs, pooled over all scenes before computing the AUC.
Pooling produces a single error distribution that weights each pair equally, which differs from a mean of per-scene AUCs.
The main text reports AUC at 1, 3, and 5 degrees, and \cref{sec:supp-full-pose} extends this to 10, 15, 20, and 30 degrees.

\subsection{Baselines}
\label{sec:supp-eval-baselines}
We reimplement each baseline into our codebase and validate it against the original code up to floating-point precision.
All methods are then evaluated under the same protocol: the same view indices, the same metric resolution, and the same resize filter.
Since a predicted scene need not align perfectly with the ground-truth cameras, we align the target cameras to the predicted scene before computing metrics for every method, following prior work~\citep{ye2024no,ye2026yonosplat}.
Pose-free methods (AnySplat~\citep{jiang2025anysplat}, EcoSplat~\citep{park2025ecosplat}, DA3~\citep{lin2025depth}, C3G~\citep{an2025c3g}, YoNoSplat~\citep{ye2026yonosplat}) predict cameras from the input images.
Posed methods (DepthSplat~\citep{xu2025depthsplat}, ZPressor~\citep{wang2026zpressor}, VolSplat~\citep{wang2026volsplat}, GlobalSplat~\citep{itkin2026globalsplat}, TokenGS~\citep{ren2026tokengs}) receive ground-truth poses and intrinsics.
Per-scene 3DGS~\citep{kerbl20233d} optimizes 100K Gaussians, initialized randomly around the camera centroids, for 30K iterations on the context views with ground-truth cameras, providing an optimization-based reference.
Each method's encoder runs at its own training resolution, and all renders are scored at a common $252{\times}252$ (\cref{sec:supp-eval-resolution}).
SparseSplat~\citep{zhang2026sparsesplat} and TokenSplat~\citep{li2026tokensplat} had no public code at the time of writing.

\subsection{Rendering resolution}
\label{sec:supp-eval-resolution}

\begin{table*}[t]
\centering
\caption{\textbf{Effect of evaluation resolution on DL3DV}.
Methods train at different native resolutions, and the scoring resolution changes the reported metrics.
\textit{Src}, \textit{Enc}, and \textit{Eval} denote the resolution for loading, encoding, and scoring.
The middle row of each baseline is our main-table protocol: each encoder runs at its native resolution, but all methods are scored at a common $252{\times}252$.
\ours{} trains at $252$, so all three settings coincide.}
\label{tbl:eval-resolution-ablation}
\scriptsize
\renewcommand{\arraystretch}{1.05}
\setlength\tabcolsep{2.5pt}
\resizebox{\linewidth}{!}{%
\begin{tabular}{lcccccccccccc}
\toprule
\multirow{2}{*}[-0.4em]{Method} & & &
& \multicolumn{3}{c}{16v}
& \multicolumn{3}{c}{32v}
& \multicolumn{3}{c}{64v} \\
\cmidrule(lr){5-7} \cmidrule(lr){8-10} \cmidrule(lr){11-13}
& Src & Enc & Eval & PSNR$\uparrow$ & SSIM$\uparrow$ & LPIPS$\downarrow$ & PSNR$\uparrow$ & SSIM$\uparrow$ & LPIPS$\downarrow$ & PSNR$\uparrow$ & SSIM$\uparrow$ & LPIPS$\downarrow$ \\
\midrule
AnySplat~\citep{jiang2025anysplat} & 540 & 448 & 448 & 18.09 & 0.556 & 0.369 & 19.42 & 0.597 & 0.338 & 19.83 & 0.613 & 0.332 \\
 & 540 & 448 & 252 & 18.40 & 0.559 & 0.314 & 19.60 & 0.603 & 0.283 & 19.91 & 0.617 & 0.278 \\
 & 252 & 252 & 252 & 14.16 & 0.296 & 0.561 & 14.74 & 0.307 & 0.564 & 14.90 & 0.317 & 0.576 \\
\midrule
YoNoSplat~\citep{ye2026yonosplat} & 540 & 224 & 224 & 20.51 & 0.647 & 0.283 & 21.03 & 0.673 & 0.262 & 21.10 & 0.682 & 0.260 \\
 & 540 & 224 & 252 & 20.36 & 0.638 & 0.299 & 20.89 & 0.664 & 0.278 & 21.00 & 0.674 & 0.275 \\
 & 252 & 252 & 252 & 18.26 & 0.492 & 0.379 & 18.38 & 0.496 & 0.378 & 18.22 & 0.495 & 0.386 \\
\midrule
\textbf{\ours} w/o TokenBA & 540 & 252 & 252 & 22.61 & 0.725 & 0.213 & 23.65 & 0.763 & 0.182 & 24.25 & 0.786 & 0.165 \\
\bottomrule
\end{tabular}}%
\end{table*}

Methods train at different resolutions and aspect ratios, and scoring at a lower resolution raises PSNR, since downsampling removes the fine detail that is hardest to reconstruct.
To compare fairly, we score every method at a common $252{\times}252$ while still running each encoder at its own training resolution.
\Cref{tbl:eval-resolution-ablation} compares this protocol with scoring each baseline at its own resolution and with running it entirely at $252$.
Scoring each baseline at its own resolution would favor the low-resolution ones and penalize the high-resolution ones: YoNoSplat gains $0.15$\,dB at its native $224$, while AnySplat loses $0.31$\,dB at its native $448$.
Running a baseline entirely at $252$ is worse still, costing AnySplat $4.2$ to $5.0$\,dB and YoNoSplat $2.1$ to $2.8$\,dB.
Our protocol therefore keeps each encoder at its training resolution, and the high-resolution baselines keep the detail they encode.

\subsection{Resize filter}
\label{sec:supp-resize-filter}

\begin{table*}[t]
\centering
\caption{\textbf{Effect of resize filter on DL3DV}.
Each method encodes at its native resolution and is scored at $252{\times}252$.
The two rows per method compare PIL Lanczos and bilinear with antialiasing as the filter for downsampling the ground-truth images.
The filter choice alone changes PSNR, so results are not comparable across papers unless the filter is reported.
We use bilinear with antialiasing throughout.}
\label{tbl:eval-resize-filter}
\scriptsize
\renewcommand{\arraystretch}{1.05}
\setlength\tabcolsep{2.5pt}
\resizebox{\linewidth}{!}{%
\begin{tabular}{lcccccccccc}
\toprule
\multirow{2}{*}[-0.4em]{Method} &
& \multicolumn{3}{c}{16v}
& \multicolumn{3}{c}{32v}
& \multicolumn{3}{c}{64v} \\
\cmidrule(lr){3-5} \cmidrule(lr){6-8} \cmidrule(lr){9-11}
& Filter & PSNR$\uparrow$ & SSIM$\uparrow$ & LPIPS$\downarrow$ & PSNR$\uparrow$ & SSIM$\uparrow$ & LPIPS$\downarrow$ & PSNR$\uparrow$ & SSIM$\uparrow$ & LPIPS$\downarrow$ \\
\midrule
AnySplat~\citep{jiang2025anysplat} & PIL Lanczos & 18.02 & 0.533 & 0.326 & 19.11 & 0.573 & 0.297 & 19.36 & 0.584 & 0.294 \\
 & bilinear$+$AA & 18.40 & 0.559 & 0.314 & 19.60 & 0.603 & 0.283 & 19.91 & 0.617 & 0.278 \\
\midrule
YoNoSplat~\citep{ye2026yonosplat} & PIL Lanczos & 19.77 & 0.602 & 0.319 & 20.24 & 0.626 & 0.299 & 20.37 & 0.638 & 0.296 \\
 & bilinear$+$AA & 20.36 & 0.638 & 0.299 & 20.89 & 0.664 & 0.278 & 21.00 & 0.674 & 0.275 \\
\midrule
\textbf{\ours} w/o TokenBA & PIL Lanczos & 21.75 & 0.689 & 0.226 & 22.71 & 0.728 & 0.195 & 23.27 & 0.752 & 0.177 \\
 & bilinear$+$AA & 22.61 & 0.725 & 0.213 & 23.65 & 0.763 & 0.182 & 24.25 & 0.786 & 0.165 \\
\bottomrule
\end{tabular}}%
\end{table*}

Switching the ground-truth downsampling filter from PIL Lanczos to bilinear with antialiasing changes PSNR by up to $0.98$\,dB (\cref{tbl:eval-resize-filter}).
Results are therefore not comparable across papers unless the filter is reported.
We use bilinear with antialiasing for every method, matching how the DL3DV source ground truth is downscaled.

\section{TokenBA Analysis}
\label{sec:supp-tokenba-analysis}
This section compares TokenBA with the test-time refinement of the baselines, breaks down its runtime, ablates its components, and shows qualitative results with and without it.

\subsection{Comparison with baseline refinement}
\label{sec:supp-ba-refinement}
\begin{table*}[t]
\centering
\caption{\textbf{Test-time refinement on DL3DV}.
Feed-forward predictions (top) and their refined counterparts (bottom), ranked separately.
TokenBA refines both Gaussians and cameras through the frozen decoder.
Gaussian-space refinement (GSR)~\citep{fan2024instantsplat} optimizes the predicted Gaussians for 60\,s per scene.
TTO is the test-time optimization of TokenGS.
Posed methods receive ground-truth cameras and have no pose metric.}
\label{tbl:supp-full-refinement}
\scriptsize
\renewcommand{\arraystretch}{1.05}
\setlength\tabcolsep{2.5pt}
\resizebox{\linewidth}{!}{%
\begin{tabular}{lcccccccccc}
\toprule
\multirow{2}{*}[-0.4em]{Method} &
& \multicolumn{3}{c}{16v}
& \multicolumn{3}{c}{32v}
& \multicolumn{3}{c}{64v} \\
\cmidrule(lr){3-5} \cmidrule(lr){6-8} \cmidrule(lr){9-11}
& Posed & PSNR$\uparrow$ & LPIPS$\downarrow$ & AUC@3$\uparrow$ & PSNR$\uparrow$ & LPIPS$\downarrow$ & AUC@3$\uparrow$ & PSNR$\uparrow$ & LPIPS$\downarrow$ & AUC@3$\uparrow$ \\
\midrule
TokenGS~\citep{ren2026tokengs} & \checkmark & \cthird 20.16 & \cthird 0.433 & N/A & \cthird 20.46 & \cthird 0.428 & N/A & \cthird 20.56 & \cthird 0.427 & N/A \\
YoNoSplat~\citep{ye2026yonosplat} &  & \csecond 20.36 & \csecond 0.299 & \csecond 0.738 & \csecond 20.89 & \csecond 0.278 & \csecond 0.755 & \csecond 21.00 & \csecond 0.275 & \csecond 0.754 \\
\textbf{\ours} w/o TokenBA &  & \cfirst 22.61 & \cfirst 0.213 & \cfirst 0.918 & \cfirst 23.65 & \cfirst 0.182 & \cfirst 0.923 & \cfirst 24.25 & \cfirst 0.165 & \cfirst 0.920 \\
\midrule
TokenGS + GSR~\citep{ren2026tokengs} & \checkmark & 21.98 & 0.272 & N/A & 24.17 & 0.214 & N/A & \cthird 25.54 & 0.185 & N/A \\
TokenGS + TTO~\citep{ren2026tokengs} & \checkmark & 21.41 & 0.345 & N/A & 22.28 & 0.332 & N/A & 22.66 & 0.328 & N/A \\
YoNoSplat + GSR~\citep{ye2026yonosplat} &  & \cthird 22.89 & \cthird 0.255 & \cthird 0.764 & \cthird 25.10 & \cthird 0.190 & \cthird 0.787 & 25.38 & \cthird 0.180 & \cthird 0.786 \\
\textbf{\ours} w/o TokenBA + GSR &  & \csecond 23.74 & \csecond 0.208 & \csecond 0.937 & \csecond 25.99 & \csecond 0.159 & \csecond 0.945 & \csecond 27.10 & \csecond 0.135 & \csecond 0.943 \\
\textbf{\ours} &  & \cfirst 24.79 & \cfirst 0.175 & \cfirst 0.954 & \cfirst 27.11 & \cfirst 0.133 & \cfirst 0.970 & \cfirst 28.99 & \cfirst 0.106 & \cfirst 0.976 \\
\bottomrule
\end{tabular}}%
\end{table*}

\Cref{tbl:supp-full-refinement} compares TokenBA with the test-time refinement available to the baselines, ranking feed-forward and refined predictions separately.
TokenGS optimizes its scene latents at test time, which lifts it by 1.3 to 2.1\,dB PSNR from 16 to 64 views.
TokenBA lifts \ours{} by 2.2 to 4.7\,dB over the same range, so the gap to TokenGS widens from 2.5 to 3.7\,dB before refinement to 3.4 to 6.3\,dB after it, although TokenGS receives ground-truth cameras.
TokenBA is also 3.2 to 3.5 times faster, taking 1.8 to 12.8\,s per scene including the feed-forward pass against 6.3 to 41.4\,s for TokenGS with its optimization.
Refining the Gaussians of \ours{} w/o TokenBA directly for 60\,s reaches 23.7 to 27.1\,dB, 1.0 to 1.9\,dB below TokenBA, which needs only 1.6 to 11.7\,s of refinement.
TokenBA also refines the cameras, raising AUC at $3^\circ$ from 0.92 to between 0.95 and 0.98, well above the 0.74 to 0.76 of YoNoSplat before refinement.
Gaussian-space refinement also helps the baselines, lifting YoNoSplat by $2.5$\,dB and TokenGS by $1.8$\,dB at 16 views, more than TokenGS gains from its own latent optimization, yet both stay $1.9$ and $2.8$\,dB below TokenBA.

\subsection{Timing breakdown}
\label{sec:supp-ba-timing}
\begin{figure*}[t]
\centering
\includegraphics[width=0.49\linewidth]{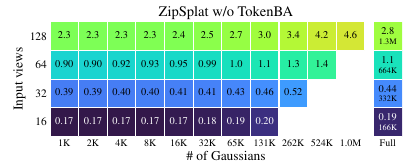}\hfill
\includegraphics[width=0.49\linewidth]{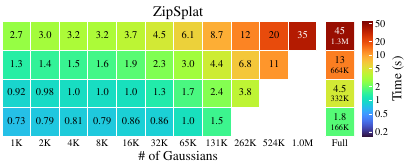}
\caption{\textbf{Inference time vs.\ input views and Gaussian budget} on DL3DV, feed-forward (left) and feed-forward + TokenBA (right), in seconds per scene on an RTX 4090.
Without TokenBA, the time depends mostly on the number of views.
TokenBA adds time that grows with both the number of views and the Gaussian count.}
\label{fig:supp-compression-time}
\end{figure*}

The cost of TokenBA is set by how much it renders.
\Cref{fig:supp-compression-time} times the grid of view counts and Gaussian budgets with and without TokenBA, and the difference between the two panels is the refinement itself.
Times follow the protocol of \cref{sec:supp-efficiency}, and the TokenBA panel averages over all 140 test scenes since its early stops make the cost scene-dependent.
This difference grows nearly linearly with both the number of views and the number of Gaussians, since every step renders all input views.
At 128 views, each doubling of the budget from 65K upward multiplies the cost of TokenBA by 1.6 to 2.0, and at 262K Gaussians each doubling of the views multiplies it by 1.7.
Without TokenBA, the three largest budgets at 128 views are up to 1.7\,s slower than full capacity.
Full capacity keeps every token and skips clustering.
A compressed budget first runs $k$-means on all 41K tokens of the 128 views.
With many clusters, each $k$-means iteration compares every token with every cluster center, and $k$-means needs 17 to 45 iterations to converge.
This costs 1.9\,s at 524K Gaussians.
\Cref{tbl:supp-ba-timing} breaks down a single step at 16 views and full capacity.
\begin{center}
\begin{minipage}{\linewidth}%
\captionsetup{type=table}
\caption{\textbf{TokenBA per-step cost} at 16 views and full capacity on an RTX 4090.}
\label{tbl:supp-ba-timing}
\footnotesize
\setlength{\tabcolsep}{4pt}
\begin{minipage}[b]{0.48\linewidth}
\centering
\begin{tabular}{lr}
\toprule
Phase & Time (ms) \\
\midrule
PnP snap           & 2.1 \\
Photometric polish & 15.0 \\
Token update       & 18.3 \\
\midrule
Total              & 35.5 \\
\bottomrule
\end{tabular}
\subcaption{One alternation round}
\end{minipage}\hfill
\begin{minipage}[b]{0.48\linewidth}
\centering
\begin{tabular}{lr}
\toprule
Phase & Time (ms) \\
\midrule
Rasterizer backward          & 10.8 \\
Rasterizer forward           & 4.3 \\
Decoder forward and backward & 2.4 \\
Adam update                  & 0.4 \\
\midrule
Total                        & 18.1 \\
\bottomrule
\end{tabular}
\subcaption{One joint tail step}
\end{minipage}
\end{minipage}
\end{center}

Each alternation round takes 36\,ms, while the rasterizer's forward and backward passes take 83\% of each tail step.
The joint tail exits after 94 of its 96 steps on average at 16 views.
Compression is therefore also the lever on the refinement cost.
At 128 views, reducing the budget from 1M to 262K Gaussians cuts TokenBA from 31\,s to 9.0\,s, at a cost of 0.8\,dB (\cref{fig:supp-compression-heatmap}).
\subsection{Ablations}
\label{sec:supp-ba-ablations}
\begin{table}[t]
\centering
\caption{\textbf{TokenBA ablations}.
Each row removes one component of the full TokenBA.
Time is per scene and includes the feed-forward pass.}
\label{tbl:supp-ba-ablations}
\scriptsize
\renewcommand{\arraystretch}{1.05}
\setlength\tabcolsep{2.5pt}
\resizebox{\linewidth}{!}{%
\begin{tabular}{lcccccccc}
\toprule
\multirow{2}{*}[-0.4em]{Method}
& \multicolumn{4}{c}{DL3DV 16v}
& \multicolumn{4}{c}{Mip-NeRF 360 32v} \\
\cmidrule(lr){2-5} \cmidrule(lr){6-9}
& PSNR$\uparrow$ & LPIPS$\downarrow$ & AUC@3$\uparrow$ & Time (s)$\downarrow$ & PSNR$\uparrow$ & LPIPS$\downarrow$ & AUC@3$\uparrow$ & Time (s)$\downarrow$ \\
\midrule
\textbf{\ours} (full TokenBA) & 24.79 & 0.175 & 0.954 & 1.80 & 24.43 & 0.210 & 0.737 & 4.76 \\
\arrayrulecolor{black!30}\cmidrule{1-9}\arrayrulecolor{black}
w/o pose update & 23.93 & 0.200 & 0.918 & 1.74 & 21.99 & 0.306 & 0.616 & 4.41 \\
w/o alternating rounds & 24.68 & 0.180 & 0.949 & 1.74 & 23.88 & 0.239 & 0.724 & 4.50 \\
w/o joint refinement & 23.16 & 0.211 & 0.944 & 0.27 & 23.16 & 0.254 & 0.685 & 0.66 \\
Joint refinement w/o poses & 24.69 & 0.178 & 0.944 & 1.80 & 24.15 & 0.218 & 0.685 & 4.70 \\
\bottomrule
\end{tabular}}%
\end{table}

\Cref{tbl:supp-ba-ablations} removes one component of TokenBA at a time, in domain on DL3DV at 16 views and out of domain on Mip-NeRF 360 at 32 views.
Updating the poses matters most.
Refitting the tokens on the predicted poses alone loses 0.9\,dB in domain and 2.4\,dB out of domain, where it even ends below TokenBA without any joint refinement.
The joint refinement carries most of the remaining gain, 1.6\,dB in domain and 1.3\,dB out of domain, but also most of the time, since the alternating rounds alone finish in 0.27\,s and 0.66\,s.
In domain, the predicted poses are already accurate, so the alternating rounds add only 0.1\,dB.
Out of domain, they recover the noisier initial poses and add 0.55\,dB.
Freezing the poses during the joint refinement costs 0.1\,dB in domain and 0.3\,dB out of domain, and the pose accuracy then stays at the value the alternating rounds reached.

\subsection{TokenBA beyond the training resolution}
\label{sec:supp-ba-resolution}
\begin{figure*}[t]
\centering
\newlength{\trw}\setlength{\trw}{\dimexpr(\linewidth - 8pt)/5\relax}%
\newcommand{\trimg}[1]{\includegraphics[width=\trw]{figures/tokenba_resolution/assets/#1.pdf}}%
\newcommand{\trrow}[1]{%
\trimg{#1_gt}\hfill\trimg{#1_ff}\hfill\trimg{#1_ba252}\hfill\trimg{#1_ba448}\hfill\trimg{#1_ba540}}%
\makebox[\trw]{\small GT}\hfill%
\makebox[\trw]{\small w/o TokenBA}\hfill%
\makebox[\trw]{\small TokenBA @ 252}\hfill%
\makebox[\trw]{\small TokenBA @ 448}\hfill%
\makebox[\trw]{\small TokenBA @ 540}%
\\[0pt]
\trrow{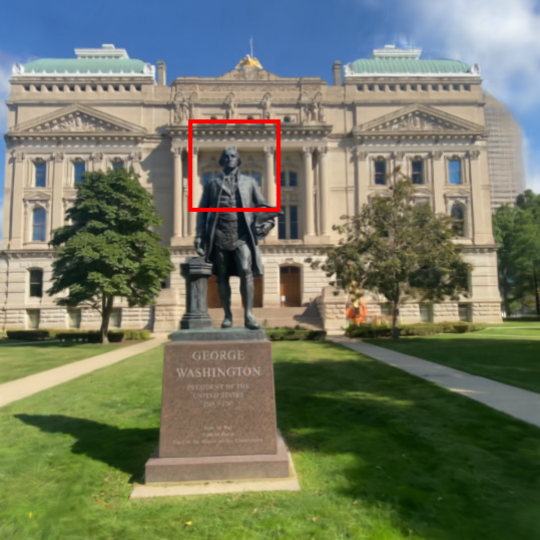}\\[1pt]
\trrow{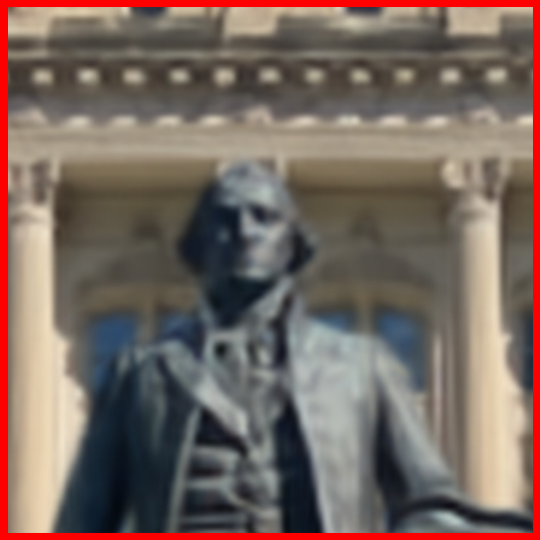}
\caption{\textbf{TokenBA at higher resolution} on DL3DV with 64 input views.
The encoder always runs at $252{\times}252$, while TokenBA fits the input views at the resolution of each column.
Fitting at a higher resolution recovers the face of the statue from the same tokens.}
\label{fig:supp-ba-resolution}
\end{figure*}

The feed-forward encoder runs at its training resolution of $252{\times}252$, but the TokenBA rendering loss is resolution-agnostic.
Refitting the tokens against higher-resolution input views recovers detail that the low-resolution pass cannot resolve.
\Cref{fig:supp-ba-resolution} shows this on DL3DV at 64 views: at $252{\times}252$, TokenBA sharpens the prediction but leaves fine features such as the statue's face blurred, whereas refitting at $448{\times}448$ or $540{\times}540$ resolves them.
The additional cost is modest -- $4.6$ times the pixels add only $1.6$ times the refinement time ($9.7$\,s to $15.6$\,s at 64 views).
All quantitative results in this paper use TokenBA at $252{\times}252$.

\subsection{Qualitative}
\label{sec:supp-ba-qualitative}
\begin{figure*}[ht]
\centering
\ifdefined\qlw\else\newlength{\qlw}\fi\setlength{\qlw}{1.1em}%
\ifdefined\qiw\else\newlength{\qiw}\fi\setlength{\qiw}{\dimexpr(\linewidth - \qlw)/6 - 2pt\relax}%
\newcommand{\bqimg}[1]{\includegraphics[width=\qiw]{figures/tokenba_qualitative/assets/#1.pdf}}%
\newcommand{\bqph}{{\setlength{\fboxsep}{0pt}\fcolorbox{black!40}{black!8}{\rule{0pt}{\dimexpr\qiw-0.8pt\relax}\rule{\dimexpr\qiw-0.8pt\relax}{0pt}}}}%
\newcommand{\bqrow}[2]{%
\begin{minipage}[c]{\qlw}\rotatebox[origin=c]{90}{\small #1}\end{minipage}%
\begin{minipage}[c]{\dimexpr\linewidth-\qlw\relax}%
\bqimg{dl3dv_016v_#2}\hfill%
\bqimg{dl3dv_032v_#2}\hfill%
\bqimg{dl3dv_064v_#2}\hfill%
\bqimg{dl3dv_128v_#2}\hfill%
\bqimg{mip_064v_#2}\hfill%
\bqimg{mip_128v_#2}%
\end{minipage}}%
\hspace{\qlw}%
\makebox[\dimexpr4\qiw+6pt\relax]{\small DL3DV}\hfill%
\makebox[\dimexpr2\qiw+2pt\relax]{\small Mip-NeRF 360}%
\\[-2pt]
\hspace{\qlw}%
\makebox[\qiw]{\small 16 views}\hfill%
\makebox[\qiw]{\small 32 views}\hfill%
\makebox[\qiw]{\small 64 views}\hfill%
\makebox[\qiw]{\small 128 views}\hfill%
\makebox[\qiw]{\small 64 views}\hfill%
\makebox[\qiw]{\small 128 views}%
\\[0pt]
\bqrow{GT}{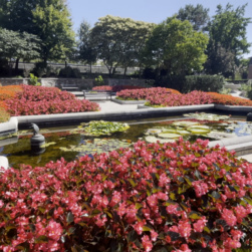}\\[2pt]
\bqrow{w/o TokenBA}{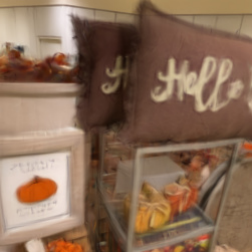}\\[2pt]
\bqrow{\ours}{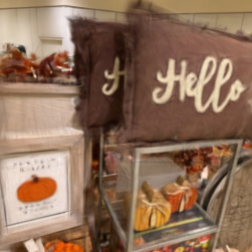}
\caption{\textbf{Effect of TokenBA.}
Target views of \ours{} without and with TokenBA against the ground truth.
The feed-forward prediction already recovers the scene well.
TokenBA sharpens fine detail such as foliage and flowers, text, and reflective surfaces, both in domain and out of domain on Mip-NeRF 360.}
\label{fig:supp-ba-qualitative}
\end{figure*}

\Cref{fig:supp-ba-qualitative} shows target views of \ours{} with and without TokenBA across view counts and out of domain.
The feed-forward prediction already reconstructs the scene well, but it averages fine appearance that the model is uncertain about.
TokenBA recovers this detail by fitting the input views through the frozen decoder, sharpening foliage and flowers, text, and reflective surfaces.

\section{Clustering Analysis}
\label{sec:supp-clustering-analysis}

This section looks at the tokens that \ours{} clusters and at the clusters it builds from them.
We use the two DA3-Base models of \cref{tbl:ablations}, trained with and without GeoCE, on DL3DV, and render without TokenBA.

\subsection{GeoCE shapes the token space}
\label{sec:supp-geoce-pca}
\begin{figure*}[t]
\centering
\includegraphics[width=\linewidth]{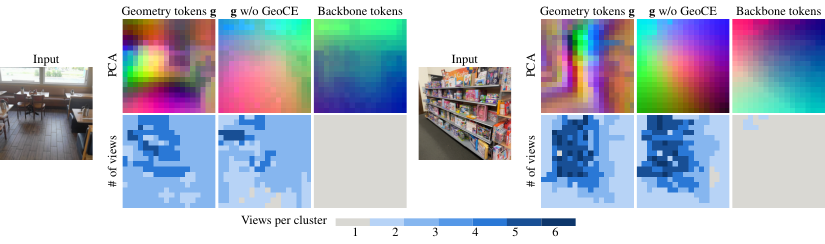}
\caption{\textbf{Token spaces and clusters.}
Top: per-patch PCA of the geometry tokens with and without GeoCE and of the pretrained backbone tokens.
Without GeoCE, the tokens vary smoothly with depth.
GeoCE separates the scene into distinct surface regions.
Bottom: number of input views spanned by each patch's cluster at $\qr{=}0.1$.
Clusters of geometry tokens merge patches across many views, whereas backbone token clusters stay within a single view.}
\label{fig:supp-geoce-pca}
\end{figure*}

GeoCE trains the geometry tokens so that patches showing the same 3D point have similar tokens, even when they come from different views.
\Cref{fig:supp-geoce-pca} shows the effect.
Without GeoCE, the tokens mostly encode depth and change smoothly across the image, and the backbone tokens behave the same way.
With GeoCE, the tokens also tell apart different objects and surfaces at the same depth.
To measure this, we take each patch and find the most similar token in the other views.
With GeoCE, this token shows the same 3D point in 88\% of cases, compared with 79\% without GeoCE and 59\% for the backbone tokens.
Even without GeoCE, the geometry tokens find the same 3D point far more often than the backbone tokens, and their clusters still span several views.
The rendering loss alone already shapes this space, since each \stokname{} is the differentiable mean of the geometry tokens it merges (\cref{sec:supp-clustering}), and the same tokens also serve as the attention keys.
Merging tokens that do not show the same surface degrades the decoded Gaussians, so training pulls corresponding tokens together.
GeoCE makes this correspondence explicit and sharper.
Clustering inherits this, as patches merged into one cluster more often show the same part of the scene, 83\% instead of 73\% at $\qr{=}0.1$.

Compression should merge patches that show the same surface in different views, since they carry the same information.
The bottom row of \cref{fig:supp-geoce-pca} shows how many views each cluster spans.
Clusters of geometry tokens span several views, whereas clusters of backbone tokens rarely leave their own view.
At $\qr{=}0.1$ and 6 views, 91\% of the clusters span at least two views, as many as when clustering the ground-truth 3D points directly, compared with 38\% for the backbone tokens.
GeoCE matters more with more views: at 32 views, a cluster spans 6.4 views with GeoCE and 5.6 without, close to the 6.7 of the ground-truth 3D points.

\subsection{Clustering method}
\label{sec:supp-cluster-space}
\begin{table}[t]
\centering
\caption{\textbf{Clustering method} on DL3DV at 6 views.
The same trained network is rendered from different groupings of its tokens.
``Nearest'' merges every token into its most similar seed without $k$-means iterations.
``Seeds only'' keeps the seed tokens and drops the rest.}
\label{tbl:supp-cluster-space}
\scriptsize
\setlength\tabcolsep{4pt}
\begin{tabular}{lllccc}
\toprule
& & & & \multicolumn{2}{c}{PSNR$\uparrow$} \\
\cmidrule(lr){5-6}
Space & Method & Init. & Merge & $\qr{=}0.25$ & $\qr{=}0.1$ \\
\midrule
Geometry tokens & $k$-means & Complexity & \checkmark & 22.92 & 21.51 \\
Geometry tokens & $k$-means & Strided & \checkmark & 22.64 & 21.38 \\
\midrule
Backbone tokens & $k$-means & Strided & \checkmark & 22.10 & 20.67 \\
Ground-truth 3D points & $k$-means & Strided & \checkmark & 22.50 & 21.26 \\
Patch color & $k$-means & Strided & \checkmark & 19.37 & 17.66 \\
Image position & Per-view grid & -- & \checkmark & 21.84 & 20.34 \\
\midrule
Geometry tokens & Nearest & Strided & \checkmark & 22.57 & 21.20 \\
Geometry tokens & Nearest & Farthest point & \checkmark & 22.59 & 21.17 \\
Geometry tokens & Nearest & Random & \checkmark & 22.48 & 21.05 \\
Geometry tokens & Seeds only & Strided &  & 20.80 & 15.72 \\
Geometry tokens & Seeds only & Random &  & 19.91 & 15.10 \\
\bottomrule
\end{tabular}
\end{table}

\ours{} compresses by running $k$-means on the geometry tokens and merging each cluster into one \stokname{}.
We test how much each part of this choice matters, namely the initialization, the space we cluster in, and the clustering itself.
\Cref{tbl:supp-cluster-space} keeps the network fixed and only changes how its tokens are grouped, and we compare each row with $k$-means on the geometry tokens under the same strided initialization.
Complexity-weighted initialization adds 0.3\,dB at $\qr{=}0.25$ and 0.1\,dB at $\qr{=}0.1$.
The space matters most.
At $\qr{=}0.1$, clustering the backbone tokens costs 0.7\,dB and the patch colors 3.7\,dB, a per-view grid on image positions costs 1.0\,dB, and even the ground-truth 3D points are 0.1\,dB worse than the geometry tokens.
The clustering algorithm matters little, since merging every token into its nearest seed without $k$-means iterations costs only 0.1 to 0.2\,dB.
Merging itself is essential, as keeping only the seed tokens loses 1.8\,dB at $\qr{=}0.25$ and 5.7\,dB at $\qr{=}0.1$.

\subsection{Compression results}
\label{sec:supp-compression}
\begin{figure*}[t]
\centering
\includegraphics[width=0.49\linewidth]{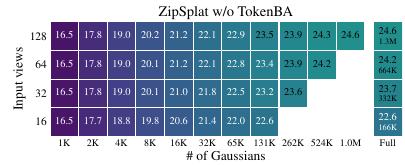}\hfill
\includegraphics[width=0.49\linewidth]{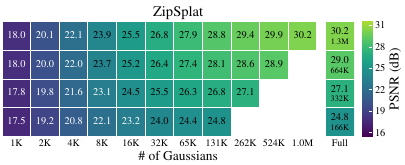}
\caption{\textbf{PSNR vs.\ input views and Gaussian budget} on DL3DV, without (left) and with (right) TokenBA.
For the same number of Gaussians, more input views give better renders.}
\label{fig:supp-compression-heatmap}
\end{figure*}

\Cref{fig:supp-compression-heatmap} shows the PSNR of \ours{} for every combination of input views and Gaussian budget.
For the same number of Gaussians, more input views give better renders.
At 131K Gaussians, going from 16 to 128 views adds 0.9\,dB without TokenBA and 4.0\,dB with it.
TokenBA helps more as the number of views and Gaussians grows, from 1.0\,dB at 16 views and 1K Gaussians to 5.7\,dB at 128 views and full capacity.
Without TokenBA, only budgets of 2K Gaussians or fewer do not improve with more views.
\Cref{fig:supp-compression-qual} shows two views at full capacity and at $4\times$ and $20\times$ compression.
What a scene loses depends on its content: large flat regions survive $20\times$ compression almost unchanged, whereas grass and thin structures need over 8K Gaussians, and TokenBA cannot fully recover them.
\begin{figure*}[t]
\centering
\ifdefined\cpw\else\newlength{\cpw}\fi%
\ifdefined\cpairgap\else\newlength{\cpairgap}\fi\setlength{\cpairgap}{5pt}%
\ifdefined\cinnergap\else\newlength{\cinnergap}\fi\setlength{\cinnergap}{2pt}%
\ifdefined\crowlabw\else\newlength{\crowlabw}\fi\setlength{\crowlabw}{11pt}%
\setlength{\cpw}{\dimexpr(\linewidth-\crowlabw-2\cpairgap-3\cinnergap-2pt)/6\relax}%
\newcommand{\cqimg}[1]{\includegraphics[width=\cpw]{figures/compression_qual/assets/#1.pdf}}%
\newcommand{\cqrow}[2]{%
  \cqimg{#1_r1_#2_rgb}\hspace{\cinnergap}\cqimg{#1_r1_#2_tok}\hspace{\cpairgap}%
  \cqimg{#1_r025_#2_rgb}\hspace{\cinnergap}\cqimg{#1_r025_#2_tok}\hspace{\cpairgap}%
  \cqimg{#1_r005_#2_rgb}\hspace{\cinnergap}\cqimg{#1_r005_#2_tok}%
}
\newcommand{\cqlab}[1]{\makebox[\dimexpr2\cpw+\cinnergap\relax][c]{\small #1}}
\newcommand{\cqrowlab}[1]{\parbox[b][\cpw][c]{\crowlabw}{\centering\rotatebox{90}{\scriptsize #1}}}
\makebox[\linewidth][c]{\cqrowlab{w/o TokenBA}\cqrow{s1}{ff}}\par\nointerlineskip\vspace{\cinnergap}%
\makebox[\linewidth][c]{\cqrowlab{with TokenBA}\cqrow{s1}{ba}}\par\nointerlineskip\vspace{\cpairgap}%
\makebox[\linewidth][c]{\cqrowlab{w/o TokenBA}\cqrow{s2}{ff}}\par\nointerlineskip\vspace{\cinnergap}%
\makebox[\linewidth][c]{\cqrowlab{with TokenBA}\cqrow{s2}{ba}}\par\nointerlineskip\vspace{2pt}%
\makebox[\linewidth][c]{\makebox[\crowlabw]{}\cqlab{full capacity}\hspace{\cpairgap}\cqlab{$4\times$ compression}\hspace{\cpairgap}\cqlab{$20\times$ compression}}
\caption{\textbf{Compression on DL3DV at 16 views.}
Two target views at full capacity, $4\times$ compression ($\qr{=}0.25$), and $20\times$ compression ($\qr{=}0.05$), each as render (left) and token map (right), without and with TokenBA.
Scenes respond differently to compression: the car scene has large flat regions, so $20\times$ compression changes the render little, whereas the grass and thin structures of the second scene need more than 8K Gaussians, and TokenBA cannot fully bring them back.}
\label{fig:supp-compression-qual}
\end{figure*}

\section{Inference Time and Rendering Speed}
\label{sec:supp-efficiency}
\begin{figure}[t]
\centering
\includegraphics[width=\linewidth]{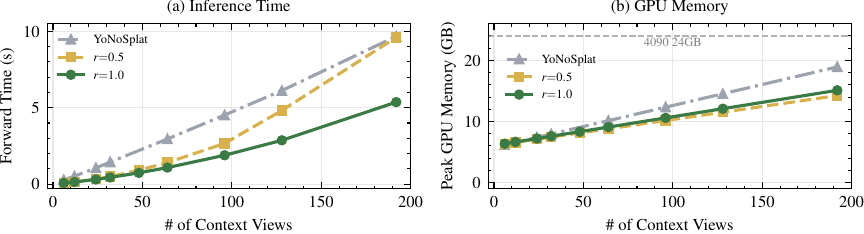}
\caption{\textbf{Forward time and memory vs.\ input views} on DL3DV.
(a) Forward time and (b) peak GPU memory of \ours{} on an RTX 4090 at full capacity and at $\qr{=}0.5$, with YoNoSplat~\citep{ye2026yonosplat} for reference.
Full capacity skips clustering, whereas $\qr{=}0.5$ pays for its $k$-means iterations at high view counts.}
\label{fig:supp-inference-cost}
\end{figure}

\begin{figure}[t]
\centering
\includegraphics[width=\linewidth]{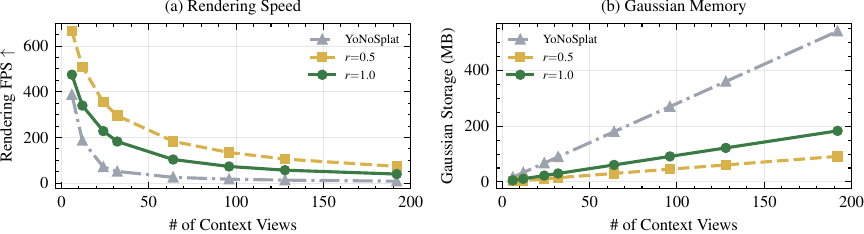}
\caption{\textbf{Rendering speed and storage vs.\ input views} on DL3DV.
(a) Rendering FPS and (b) storage of the predicted Gaussians at full capacity and at $\qr{=}0.5$, with YoNoSplat~\citep{ye2026yonosplat} for reference.
The Gaussian count grows linearly with the views, and $\qr{=}0.5$ halves the storage and nearly doubles the rendering speed.}
\label{fig:supp-gaussian-cost}
\end{figure}

\paragraph{Setup.}
All measurements run on a single NVIDIA RTX 4090 (24\,GB) at $252{\times}252$ resolution with batch size 1, with the backbone and fusion in bf16.
Each configuration averages 15 timed forward passes after 5 warm-up passes on a single DL3DV scene.
The forward pass covers the prediction of the Gaussians and excludes data loading, TokenBA and the evaluation renders.
We time full capacity ($\qr{=}1.0$), which keeps every token and skips clustering, and a fixed ratio of $\qr{=}0.5$.

\paragraph{Inference cost.}
Forward time grows with the number of input views (\cref{fig:supp-inference-cost}a), from 0.07\,s at 6 views to 5.4\,s at 192 views at full capacity.
Full capacity is as fast as the compressed ratios because it skips clustering, and the rest of the network is shallow.
The 40-layer backbone takes about three quarters of the time, and the three fusion blocks, the color attention and the Gaussian head add the remaining quarter even though every token is a query.
Compression can only shrink this last quarter, and it first pays for $k$-means.
At $\qr{=}0.5$ this costs little up to 32 views, but the $k$-means iterations grow with the number of tokens times the number of clusters.
At 192 views, $\qr{=}0.5$ takes 9.6\,s, the same as YoNoSplat, and the 4.2\,s it adds over full capacity are the $k$-means iterations, which compare all 62K tokens with 31K cluster centers in every iteration (\cref{sec:supp-ba-timing}).
Peak GPU memory (\cref{fig:supp-inference-cost}b) grows from 6.4\,GB at 6 views to 15.1\,GB at 192 views at full capacity, and $\qr{=}0.5$ needs slightly less.
YoNoSplat reaches 18.9\,GB at 192 views.

\paragraph{Output representation.}
The Gaussian count sets the cost of the final representation (\cref{fig:supp-gaussian-cost}).
At full capacity it grows linearly with the number of input views, from 62K Gaussians at 6 views to 2.0M at 192 views (183\,MB, 40\,FPS at $252{\times}252$).
Halving the ratio halves the storage and nearly doubles the rendering speed, to 92\,MB and 74\,FPS at 192 views.
At 24 views, full capacity renders 249K Gaussians at 228\,FPS and $\qr{=}0.5$ renders 124K at 354\,FPS, well within the limits of real-time viewers.
YoNoSplat produces 9.6M Gaussians at 192 views (539\,MB) and renders at 8.9\,FPS, about $5{\times}$ more primitives and $4.5{\times}$ slower than \ours{} at full capacity.

\section{More Results}
\label{sec:supp-full-results}
The tables in this section extend the main results to every baseline, every view count, and every variant of \ours{}.
\Cref{tbl:supp-full-dl3dv} covers the DL3DV ladder at 16, 32, 64, and 128 views, and \cref{tbl:supp-full-re10k,tbl:supp-full-mipnerf360,tbl:supp-full-scannetpp} the remaining benchmarks.
Each table lists \ours{} without TokenBA, at fixed ratios $\qr{=}0.5$ and $0.1$ or at the fixed budget of 262K Gaussians of TokenGS (with TokenBA), and with ground-truth cameras (posed).

\subsection{DL3DV}
\label{sec:supp-full-dl3dv}
\begin{table*}[t]
\centering
\caption{\textbf{Full DL3DV results}.
All methods at 16, 32, 64, and 128 input views.
Posed methods receive ground-truth cameras.
\#Gs is the total number of predicted Gaussians.}
\label{tbl:supp-full-dl3dv}
\scriptsize
\renewcommand{\arraystretch}{1.05}
\setlength\tabcolsep{2.5pt}
\resizebox{\linewidth}{!}{%
\begin{tabular}{lccccccccccccccccc}
\toprule
\multirow{2}{*}[-0.4em]{Method} &
& \multicolumn{4}{c}{16v}
& \multicolumn{4}{c}{32v}
& \multicolumn{4}{c}{64v}
& \multicolumn{4}{c}{128v} \\
\cmidrule(lr){3-6} \cmidrule(lr){7-10} \cmidrule(lr){11-14} \cmidrule(lr){15-18}
& Posed & \#Gs & PSNR$\uparrow$ & SSIM$\uparrow$ & LPIPS$\downarrow$ & \#Gs & PSNR$\uparrow$ & SSIM$\uparrow$ & LPIPS$\downarrow$ & \#Gs & PSNR$\uparrow$ & SSIM$\uparrow$ & LPIPS$\downarrow$ & \#Gs & PSNR$\uparrow$ & SSIM$\uparrow$ & LPIPS$\downarrow$ \\
\midrule
DepthSplat~\citep{xu2025depthsplat} & \checkmark & 1.8M & 19.64 & 0.660 & 0.278 & 3.7M & 18.61 & 0.623 & 0.318 & 7.3M & 17.32 & 0.555 & 0.390 & 14.7M & 15.98 & 0.461 & 0.489 \\
ZPressor~\citep{wang2026zpressor} & \checkmark & 688K & 18.44 & 0.575 & 0.339 & 688K & 18.34 & 0.569 & 0.344 & 688K & 18.36 & 0.568 & 0.345 & 688K & 18.44 & 0.570 & 0.344 \\
VolSplat~\citep{wang2026volsplat} & \checkmark & 1.0M & 15.42 & 0.385 & 0.565 & 2.1M & 15.64 & 0.399 & 0.560 & 4.2M & 15.55 & 0.398 & 0.566 & \multicolumn{4}{c}{OOM} \\
GlobalSplat~\citep{itkin2026globalsplat} & \checkmark & 16K & 15.48 & 0.397 & 0.630 & 16K & 15.68 & 0.402 & 0.625 & 16K & 15.72 & 0.402 & 0.626 & 16K & 15.80 & 0.404 & 0.628 \\
TokenGS w/o TTO~\citep{ren2026tokengs} & \checkmark & 262K & 20.16 & 0.614 & 0.433 & 262K & 20.46 & 0.626 & 0.428 & 262K & 20.56 & 0.631 & 0.427 & 262K & 20.62 & 0.634 & 0.425 \\
TokenGS~\citep{ren2026tokengs} & \checkmark & 262K & 21.41 & 0.695 & 0.345 & 262K & 22.28 & 0.727 & 0.332 & 262K & 22.66 & 0.740 & 0.328 & 262K & 22.86 & 0.748 & 0.324 \\
\arrayrulecolor{black!30}\cmidrule{1-18}\arrayrulecolor{black}
AnySplat~\citep{jiang2025anysplat} &  & 2.3M & 18.40 & 0.559 & 0.314 & 4.1M & 19.60 & 0.603 & 0.283 & 7.2M & 19.91 & 0.617 & 0.278 & 12.3M & 19.95 & 0.622 & 0.279 \\
EcoSplat ($r{=}0.4$)~\citep{park2025ecosplat} &  & 419K & 14.05 & 0.319 & 0.605 & 839K & 13.69 & 0.302 & 0.626 & 1.7M & 13.06 & 0.275 & 0.647 & 3.4M & 12.84 & 0.268 & 0.661 \\
EcoSplat ($r{=}0.95$)~\citep{park2025ecosplat} &  & 971K & 13.84 & 0.305 & 0.611 & 1.9M & 13.44 & 0.288 & 0.632 & 3.9M & 12.78 & 0.262 & 0.651 & 7.8M & 12.61 & 0.258 & 0.662 \\
C3G~\citep{an2025c3g} &  & 2K & 12.71 & 0.286 & 0.649 & 2K & 11.97 & 0.258 & 0.661 & 2K & 11.56 & 0.243 & 0.672 & 2K & 11.51 & 0.244 & 0.678 \\
DA3~\citep{lin2025depth} &  & 4.1M & 20.34 & 0.652 & 0.276 & 8.1M & 20.82 & 0.673 & 0.255 & 16.3M & 20.59 & 0.666 & 0.257 & 32.5M & 20.09 & 0.646 & 0.271 \\
YoNoSplat~\citep{ye2026yonosplat} &  & 803K & 20.36 & 0.638 & 0.299 & 1.6M & 20.89 & 0.664 & 0.278 & 3.2M & 21.00 & 0.674 & 0.275 & 6.4M & 20.42 & 0.661 & 0.293 \\
\arrayrulecolor{black!30}\cmidrule{1-18}\arrayrulecolor{black}
\textbf{\ours} w/o TokenBA &  & 166K & 22.61 & 0.725 & 0.213 & 332K & 23.65 & 0.763 & 0.182 & 664K & 24.25 & 0.786 & 0.165 & 1.3M & 24.56 & 0.800 & 0.155 \\
\textbf{\ours} ($\qr{=}0.1$) &  & 17K & 23.22 & 0.739 & 0.293 & 33K & 25.51 & 0.809 & 0.229 & 66K & 27.39 & 0.857 & 0.179 & 133K & 28.76 & 0.887 & 0.145 \\
\textbf{\ours} ($\qr{=}0.5$) &  & 83K & \cthird 24.57 & \cthird 0.798 & \cthird 0.200 & 166K & \cthird 26.89 & \cthird 0.856 & \cthird 0.153 & 332K & \cthird 28.74 & \cthird 0.893 & \cthird 0.122 & 664K & \cthird 29.98 & \cthird 0.914 & \cthird 0.102 \\
\textbf{\ours} &  & 166K & \csecond 24.79 & \csecond 0.808 & \csecond 0.175 & 332K & \csecond 27.11 & \csecond 0.865 & \csecond 0.133 & 664K & \csecond 28.99 & \csecond 0.900 & \csecond 0.106 & 1.3M & \csecond 30.22 & \csecond 0.920 & \csecond 0.090 \\
\textbf{\ours} & \checkmark & 166K & \cfirst 24.88 & \cfirst 0.813 & \cfirst 0.173 & 332K & \cfirst 27.23 & \cfirst 0.870 & \cfirst 0.130 & 664K & \cfirst 29.14 & \cfirst 0.905 & \cfirst 0.104 & 1.3M & \cfirst 30.41 & \cfirst 0.926 & \cfirst 0.088 \\
\bottomrule
\end{tabular}}%
\end{table*}

No baseline exceeds 23\,dB at any view count, and no pose-free baseline exceeds 21.0\,dB.
DepthSplat, DA3, and YoNoSplat decline beyond 32 or 64 views, and the fixed-budget methods gain at most 1.45\,dB (TokenGS) across the ladder, whereas \ours{} gains 1.95\,dB from 16 to 128 views without TokenBA and 5.43\,dB with it.
A fixed ratio of $\qr{=}0.5$ stays within 0.25\,dB of full capacity with half the Gaussians, and $\qr{=}0.1$ stays within 1.6\,dB with a tenth while remaining at least 1.8\,dB above every baseline.
Ground-truth cameras add at most 0.19\,dB, so the predicted cameras cost almost nothing in domain.

\subsection{RealEstate10K}
\label{sec:supp-full-re10k}
\begin{table*}[t]
\centering
\caption{\textbf{Full RealEstate10K results} at 6 input views.
Posed methods receive ground-truth cameras; NoPoSplat receives only intrinsics.
\#Gs is the total number of predicted Gaussians.}
\label{tbl:supp-full-re10k}
\scriptsize
\renewcommand{\arraystretch}{1.05}
\setlength\tabcolsep{2.5pt}
\begin{tabular}{lccccc}
\toprule
\multirow{2}{*}[-0.4em]{Method} &
& \multicolumn{4}{c}{RE10K 6v} \\
\cmidrule(lr){3-6}
& Posed & \#Gs & PSNR$\uparrow$ & SSIM$\uparrow$ & LPIPS$\downarrow$ \\
\midrule
DepthSplat~\citep{xu2025depthsplat} & \checkmark & 688K & 24.62 & 0.873 & 0.138 \\
ZPressor~\citep{wang2026zpressor} & \checkmark & 688K & 24.89 & 0.864 & 0.141 \\
VolSplat~\citep{wang2026volsplat} & \checkmark & 393K & 22.98 & 0.807 & 0.230 \\
GlobalSplat~\citep{itkin2026globalsplat} & \checkmark & 16K & 23.68 & 0.782 & 0.232 \\
TokenGS w/o TTO~\citep{ren2026tokengs} & \checkmark & 262K & 26.57 & 0.865 & 0.199 \\
TokenGS~\citep{ren2026tokengs} & \checkmark & 262K & 27.59 & \csecond 0.899 & 0.145 \\
\arrayrulecolor{black!30}\cmidrule{1-6}\arrayrulecolor{black}
NoPoSplat~\citep{ye2024no} &  & 393K & 20.22 & 0.680 & 0.311 \\
AnySplat~\citep{jiang2025anysplat} &  & 775K & 22.60 & 0.791 & 0.189 \\
EcoSplat ($r{=}0.4$)~\citep{park2025ecosplat} &  & 157K & 21.73 & 0.728 & 0.267 \\
EcoSplat ($r{=}0.95$)~\citep{park2025ecosplat} &  & 362K & 21.82 & 0.733 & 0.265 \\
C3G~\citep{an2025c3g} &  & 2K & 20.64 & 0.663 & 0.318 \\
DA3~\citep{lin2025depth} &  & 1.5M & 21.37 & 0.726 & 0.229 \\
YoNoSplat~\citep{ye2026yonosplat} &  & 301K & 25.05 & 0.830 & 0.154 \\
\arrayrulecolor{black!30}\cmidrule{1-6}\arrayrulecolor{black}
\textbf{\ours} w/o TokenBA &  & 62K & 26.86 & 0.856 & 0.134 \\
\textbf{\ours} ($\qr{=}0.1$) &  & 6K & 26.76 & 0.846 & 0.198 \\
\textbf{\ours} ($\qr{=}0.5$) &  & 31K & \cthird 28.34 & 0.880 & \cthird 0.130 \\
\textbf{\ours} &  & 62K & \csecond 28.50 & \cthird 0.882 & \csecond 0.120 \\
\textbf{\ours} & \checkmark & 62K & \cfirst 30.13 & \cfirst 0.924 & \cfirst 0.102 \\
\bottomrule
\end{tabular}
\end{table*}

Without TokenBA, \ours{} already exceeds TokenGS w/o TTO and YoNoSplat with 62K Gaussians, four and five times fewer.
At $\qr{=}0.1$, 6K Gaussians reach the PSNR of the full feed-forward model.
Ground-truth cameras add 1.63\,dB here against 0.1\,dB on DL3DV, in line with the lower pose accuracy at six views (\cref{tbl:supp-full-pose}).

\subsection{Mip-NeRF360}
\label{sec:supp-full-mipnerf360}
\begin{table*}[t]
\centering
\caption{\textbf{Full Mip-NeRF 360 results} at 32, 64, and 128 input views.
No method is trained on this dataset.
Posed methods receive ground-truth cameras.
\#Gs is the total number of predicted Gaussians.
The per-scene 3DGS fit is a grey reference excluded from the ranking.}
\label{tbl:supp-full-mipnerf360}
\scriptsize
\renewcommand{\arraystretch}{1.05}
\setlength\tabcolsep{2.5pt}
\resizebox{\linewidth}{!}{%
\begin{tabular}{lccccccccccccc}
\toprule
\multirow{2}{*}[-0.4em]{Method} &
& \multicolumn{4}{c}{32v}
& \multicolumn{4}{c}{64v}
& \multicolumn{4}{c}{128v} \\
\cmidrule(lr){3-6} \cmidrule(lr){7-10} \cmidrule(lr){11-14}
& Posed & \#Gs & PSNR$\uparrow$ & SSIM$\uparrow$ & LPIPS$\downarrow$ & \#Gs & PSNR$\uparrow$ & SSIM$\uparrow$ & LPIPS$\downarrow$ & \#Gs & PSNR$\uparrow$ & SSIM$\uparrow$ & LPIPS$\downarrow$ \\
\midrule
\color{black!50}3DGS$_{\text{30k}}$~\citep{kerbl20233d} & \color{black!50}\checkmark & \color{black!50}1.0M & \color{black!50}21.28 & \color{black!50}0.678 & \color{black!50}0.282 & \color{black!50}1.2M & \color{black!50}24.84 & \color{black!50}0.806 & \color{black!50}0.188 & \color{black!50}1.3M & \color{black!50}27.21 & \color{black!50}0.862 & \color{black!50}0.137 \\
\arrayrulecolor{black!30}\cmidrule{1-14}\arrayrulecolor{black}
DepthSplat~\citep{xu2025depthsplat} & \checkmark & 3.7M & 16.33 & 0.443 & 0.455 & 7.3M & 15.62 & 0.405 & 0.499 & 14.7M & 14.65 & 0.356 & 0.585 \\
ZPressor~\citep{wang2026zpressor} & \checkmark & 688K & 17.29 & 0.445 & 0.446 & 688K & 17.34 & 0.447 & 0.448 & 688K & 16.93 & 0.418 & 0.464 \\
GlobalSplat~\citep{itkin2026globalsplat} & \checkmark & 16K & 14.91 & 0.375 & 0.753 & 16K & 14.56 & 0.360 & 0.748 & 16K & 15.14 & 0.380 & 0.747 \\
TokenGS w/o TTO~\citep{ren2026tokengs} & \checkmark & 262K & 19.77 & 0.507 & 0.569 & 262K & 19.76 & 0.510 & 0.564 & 262K & 19.92 & 0.517 & 0.561 \\
TokenGS~\citep{ren2026tokengs} & \checkmark & 262K & 22.84 & 0.648 & 0.434 & 262K & 23.07 & 0.656 & 0.432 & 262K & 23.22 & 0.663 & 0.424 \\
\arrayrulecolor{black!30}\cmidrule{1-14}\arrayrulecolor{black}
AnySplat~\citep{jiang2025anysplat} &  & 4.5M & 19.10 & 0.537 & 0.322 & 8.2M & 19.91 & 0.551 & 0.302 & 14.5M & 20.03 & 0.556 & 0.302 \\
EcoSplat~\citep{park2025ecosplat} &  & 839K & 14.98 & 0.320 & 0.650 & 1.7M & 14.85 & 0.308 & 0.657 & 3.4M & 14.36 & 0.297 & 0.681 \\
C3G~\citep{an2025c3g} &  & 2K & 14.59 & 0.315 & 0.690 & 2K & 14.48 & 0.318 & 0.693 & 2K & 14.28 & 0.316 & 0.702 \\
DA3~\citep{lin2025depth} &  & 8.1M & 21.48 & 0.603 & 0.266 & 16.3M & 21.03 & 0.585 & 0.277 & 32.5M & 20.62 & 0.579 & 0.287 \\
YoNoSplat~\citep{ye2026yonosplat} &  & 1.6M & 17.61 & 0.411 & 0.484 & 3.2M & 17.73 & 0.417 & 0.479 & 6.4M & 17.54 & 0.416 & 0.494 \\
\arrayrulecolor{black!30}\cmidrule{1-14}\arrayrulecolor{black}
\textbf{\ours} w/o TokenBA &  & 332K & 22.30 & 0.606 & 0.256 & 664K & 22.53 & 0.620 & 0.244 & 1.3M & 22.81 & 0.634 & 0.231 \\
\textbf{\ours} (262K) &  & 262K & \cthird 24.37 & \cthird 0.695 & \cthird 0.216 & 262K & \cthird 25.16 & \cthird 0.719 & \cthird 0.217 & 262K & \cthird 25.66 & \cthird 0.733 & \cthird 0.216 \\
\textbf{\ours} &  & 332K & \csecond 24.43 & \csecond 0.696 & \csecond 0.210 & 664K & \csecond 25.24 & \csecond 0.723 & \csecond 0.185 & 1.3M & \csecond 25.81 & \csecond 0.739 & \csecond 0.169 \\
\textbf{\ours} & \checkmark & 332K & \cfirst 26.57 & \cfirst 0.820 & \cfirst 0.148 & 664K & \cfirst 27.38 & \cfirst 0.845 & \cfirst 0.125 & 1.3M & \cfirst 28.35 & \cfirst 0.870 & \cfirst 0.106 \\
\bottomrule
\end{tabular}}%
\end{table*}

Without TokenBA, \ours{} exceeds every feed-forward baseline at every view count, and its PSNR rises from 32 to 128 views while DA3, which shares our backbone, declines by 0.86\,dB.
At the fixed budget of 262K Gaussians, matching TokenGS, \ours{} stays within 0.15\,dB of full capacity at every view count, with five times fewer Gaussians at 128 views.
With ground-truth cameras, \ours{} gains 2.1 to 2.5\,dB and exceeds the per-scene 3DGS fit at every view count, so out of domain the gap to oracle cameras is ten times larger than on DL3DV.

\subsection{ScanNet++}
\label{sec:supp-full-scannetpp}
\begin{table*}[t]
\centering
\caption{\textbf{Full ScanNet++ results} on iPhone captures at 32, 64, and 128 input views.
Posed methods receive ground-truth cameras.
\#Gs is the total number of predicted Gaussians.
The per-scene 3DGS fit is a grey reference excluded from the ranking.
*: AnySplat and DA3 are trained on ScanNet++.}
\label{tbl:supp-full-scannetpp}
\scriptsize
\renewcommand{\arraystretch}{1.05}
\setlength\tabcolsep{2.5pt}
\resizebox{\linewidth}{!}{%
\begin{tabular}{lccccccccccccc}
\toprule
\multirow{2}{*}[-0.4em]{Method} &
& \multicolumn{4}{c}{32v}
& \multicolumn{4}{c}{64v}
& \multicolumn{4}{c}{128v} \\
\cmidrule(lr){3-6} \cmidrule(lr){7-10} \cmidrule(lr){11-14}
& Posed & \#Gs & PSNR$\uparrow$ & SSIM$\uparrow$ & LPIPS$\downarrow$ & \#Gs & PSNR$\uparrow$ & SSIM$\uparrow$ & LPIPS$\downarrow$ & \#Gs & PSNR$\uparrow$ & SSIM$\uparrow$ & LPIPS$\downarrow$ \\
\midrule
\color{black!50}3DGS$_{\text{30k}}$~\citep{kerbl20233d} & \color{black!50}\checkmark & \color{black!50}1.2M & \color{black!50}20.11 & \color{black!50}0.690 & \color{black!50}0.365 & \color{black!50}1.6M & \color{black!50}21.16 & \color{black!50}0.732 & \color{black!50}0.345 & \color{black!50}1.7M & \color{black!50}22.14 & \color{black!50}0.775 & \color{black!50}0.324 \\
\arrayrulecolor{black!30}\cmidrule{1-14}\arrayrulecolor{black}
DepthSplat~\citep{xu2025depthsplat} & \checkmark & 3.7M & 17.96 & 0.705 & 0.393 & 7.3M & 17.35 & 0.688 & 0.422 & 14.7M & 16.65 & 0.660 & 0.469 \\
ZPressor~\citep{wang2026zpressor} & \checkmark & 688K & 15.42 & 0.575 & 0.468 & 688K & 16.56 & 0.627 & 0.432 & 688K & 16.54 & 0.628 & 0.432 \\
GlobalSplat~\citep{itkin2026globalsplat} & \checkmark & 16K & 14.80 & 0.606 & 0.633 & 16K & 14.93 & 0.612 & 0.632 & 16K & 14.88 & 0.611 & 0.636 \\
TokenGS w/o TTO~\citep{ren2026tokengs} & \checkmark & 262K & 19.80 & 0.714 & 0.505 & 262K & 19.64 & 0.711 & 0.513 & 262K & 19.59 & 0.712 & 0.515 \\
TokenGS~\citep{ren2026tokengs} & \checkmark & 262K & 22.14 & 0.772 & 0.422 & 262K & 22.49 & 0.782 & 0.426 & 262K & 22.71 & 0.786 & 0.427 \\
\arrayrulecolor{black!30}\cmidrule{1-14}\arrayrulecolor{black}
AnySplat*~\citep{jiang2025anysplat} &  & 3.5M & 16.63 & 0.669 & 0.391 & 6.9M & 18.36 & 0.694 & 0.348 & 12.8M & 18.53 & 0.692 & 0.354 \\
EcoSplat~\citep{park2025ecosplat} &  & 839K & 14.07 & 0.515 & 0.642 & 1.7M & 13.67 & 0.498 & 0.654 & 3.4M & 13.75 & 0.506 & 0.667 \\
C3G~\citep{an2025c3g} &  & 2K & 12.24 & 0.487 & 0.627 & 2K & 12.00 & 0.479 & 0.635 & 2K & 11.71 & 0.468 & 0.641 \\
DA3*~\citep{lin2025depth} &  & 8.1M & 18.48 & 0.683 & 0.394 & 16.3M & 19.15 & 0.702 & 0.362 & 32.5M & 19.12 & 0.703 & 0.361 \\
YoNoSplat~\citep{ye2026yonosplat} &  & 1.6M & 16.13 & 0.622 & 0.516 & 3.2M & 16.29 & 0.626 & 0.506 & 6.4M & 16.12 & 0.623 & 0.513 \\
\arrayrulecolor{black!30}\cmidrule{1-14}\arrayrulecolor{black}
\textbf{\ours} w/o TokenBA &  & 332K & 20.34 & 0.729 & 0.324 & 664K & 21.42 & 0.758 & 0.278 & 1.3M & 21.76 & 0.770 & 0.265 \\
\textbf{\ours} (262K) &  & 262K & \csecond 22.46 & \csecond 0.780 & \cthird 0.299 & 262K & \cthird 24.29 & \cthird 0.821 & \cthird 0.263 & 262K & \cthird 25.23 & \cthird 0.842 & \cthird 0.260 \\
\textbf{\ours} &  & 332K & \cthird 22.36 & \cthird 0.779 & \csecond 0.298 & 664K & \csecond 24.36 & \csecond 0.826 & \csecond 0.249 & 1.3M & \csecond 25.40 & \csecond 0.848 & \csecond 0.235 \\
\textbf{\ours} & \checkmark & 332K & \cfirst 25.35 & \cfirst 0.863 & \cfirst 0.209 & 664K & \cfirst 26.78 & \cfirst 0.891 & \cfirst 0.186 & 1.3M & \cfirst 27.93 & \cfirst 0.905 & \cfirst 0.176 \\
\bottomrule
\end{tabular}}%
\end{table*}

\ours{} without cameras exceeds the per-scene 3DGS fit at every view count, by 2.25 to 3.26\,dB.
Without TokenBA it also exceeds AnySplat and DA3, although both train on ScanNet++.
Ground-truth cameras add 2.99 and 2.42\,dB at 32 and 64 views.

\subsection{Pose accuracy}
\label{sec:supp-full-pose}
\begin{table*}[t]
\centering
\caption{\textbf{Full pose accuracy}.
AUC of the relative rotation and translation-direction error over all input-view pairs, pooled over all scenes, at thresholds from $1^\circ$ to $30^\circ$.
DL3DV uses 64 input views; RealEstate10K uses 6.}
\label{tbl:supp-full-pose}
\scriptsize
\renewcommand{\arraystretch}{1.05}
\setlength\tabcolsep{2.5pt}
\resizebox{\linewidth}{!}{%
\begin{tabular}{lcccccccccccccc}
\toprule
Method
& \multicolumn{7}{c}{DL3DV 64v}
& \multicolumn{7}{c}{RE10K 6v} \\
\cmidrule(lr){2-8} \cmidrule(lr){9-15}
\multicolumn{1}{r}{AUC$\uparrow$} & $1^\circ$ & $3^\circ$ & $5^\circ$ & $10^\circ$ & $15^\circ$ & $20^\circ$ & $30^\circ$ & $1^\circ$ & $3^\circ$ & $5^\circ$ & $10^\circ$ & $15^\circ$ & $20^\circ$ & $30^\circ$ \\
\midrule
AnySplat~\citep{jiang2025anysplat} & 0.071 & 0.357 & 0.550 & 0.755 & 0.833 & 0.873 & 0.914 & 0.130 & 0.378 & 0.530 & 0.715 & 0.795 & 0.839 & 0.886 \\
DA3~\citep{lin2025depth} & \csecond 0.930 & \csecond 0.969 & \csecond 0.980 & \csecond 0.989 & \cfirst 0.993 & \cfirst 0.994 & \cfirst 0.996 & \cthird 0.546 & \cthird 0.741 & \cthird 0.817 & \cthird 0.890 & \cthird 0.920 & \cthird 0.935 & \cthird 0.953 \\
YoNoSplat~\citep{ye2026yonosplat} & 0.466 & 0.754 & 0.842 & 0.915 & \cthird 0.941 & \cthird 0.955 & \cthird 0.969 & 0.361 & 0.673 & 0.779 & 0.876 & 0.913 & 0.933 & \cthird 0.953 \\
\arrayrulecolor{black!30}\cmidrule{1-15}\arrayrulecolor{black}
\textbf{\ours} w/o TokenBA & \cthird 0.827 & \cthird 0.920 & \cthird 0.948 & \cthird 0.972 & \csecond 0.980 & \csecond 0.985 & \csecond 0.989 & \csecond 0.706 & \csecond 0.843 & \csecond 0.890 & \csecond 0.936 & \csecond 0.954 & \csecond 0.964 & \csecond 0.975 \\
\textbf{\ours} & \cfirst 0.950 & \cfirst 0.976 & \cfirst 0.983 & \cfirst 0.990 & \cfirst 0.993 & \cfirst 0.994 & \cfirst 0.996 & \cfirst 0.738 & \cfirst 0.855 & \cfirst 0.897 & \cfirst 0.939 & \cfirst 0.956 & \cfirst 0.965 & \cfirst 0.976 \\
\bottomrule
\end{tabular}}%
\end{table*}

With TokenBA, \ours{} is the most accurate at every threshold up to $10^\circ$ on both datasets and ties DA3 from $15^\circ$ on DL3DV.
Our feed-forward camera head is initialized from DA3's.
At 64 views on DL3DV it trails DA3 at $1^\circ$ (0.827 against 0.930), while at six views on RealEstate10K it is far ahead (0.706 against 0.546).
TokenBA lifts it above DA3 in both cases.

\clearpage
\renewcommand{\topfraction}{0.99}
\renewcommand{\dbltopfraction}{0.99}
\renewcommand{\textfraction}{0.01}
\renewcommand{\floatpagefraction}{0.95}

\section{Additional Qualitative Results}
\label{sec:supp-qualitative}

\Cref{fig:supp-qual-dl3dv-more} extends \cref{fig:qualitative} with more DL3DV scenes at 16, 32, 64, and 128 input views.
We show additional qualitative comparisons on RealEstate10K in \cref{fig:supp-qual-re10k}.

\begin{figure*}[ht]
\centering
\ifdefined\qlw\else\newlength{\qlw}\fi\setlength{\qlw}{1.1em}%
\ifdefined\qiw\else\newlength{\qiw}\fi\setlength{\qiw}{\dimexpr(\linewidth - \qlw)/6 - 2pt\relax}%
\newcommand{\qimg}[1]{\includegraphics[width=\qiw]{figures/qualitative/assets/#1}}%
\hspace{\qlw}%
\makebox[\qiw]{\small GT}\hfill%
\makebox[\qiw]{\small \ours}\hfill%
\makebox[\qiw]{\small \ours{} w/o tBA}\hfill%
\makebox[\qiw]{\small TokenGS}\hfill%
\makebox[\qiw]{\small YoNoSplat}\hfill%
\makebox[\qiw]{\small DA3}%
\\[0pt]
\begin{minipage}[c]{\qlw}%
\rotatebox[origin=c]{90}{\small 16 views}%
\end{minipage}%
\begin{minipage}[c]{\dimexpr\linewidth-\qlw\relax}%
\qimg{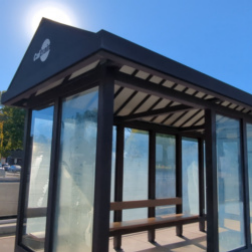}\hfill%
\qimg{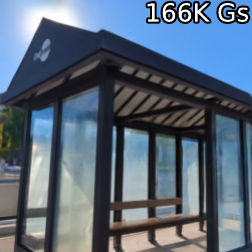}\hfill%
\qimg{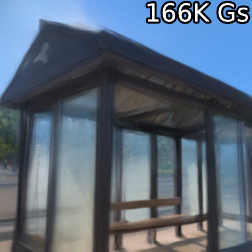}\hfill%
\qimg{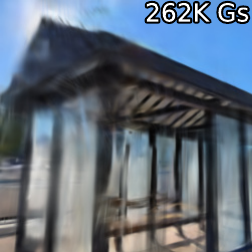}\hfill%
\qimg{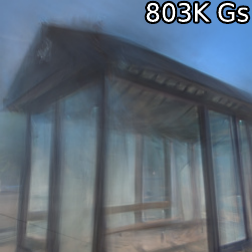}\hfill%
\qimg{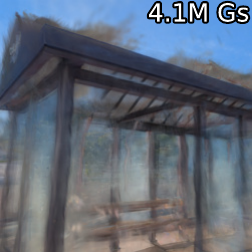}%
\\[2pt]
\qimg{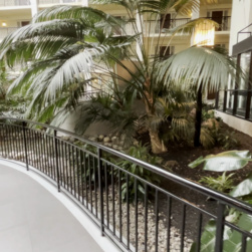}\hfill%
\qimg{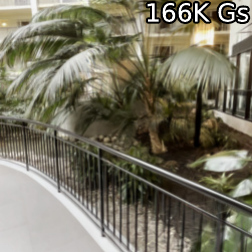}\hfill%
\qimg{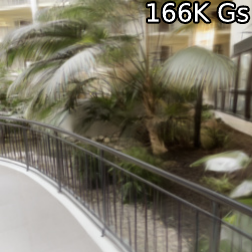}\hfill%
\qimg{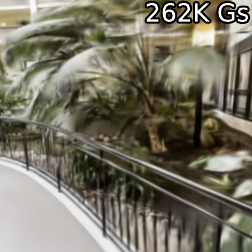}\hfill%
\qimg{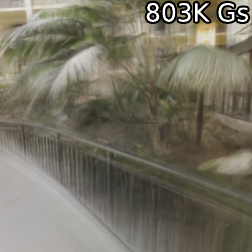}\hfill%
\qimg{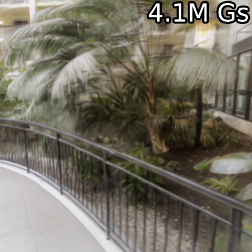}%
\end{minipage}%
\\[2pt]
\begin{minipage}[c]{\qlw}%
\rotatebox[origin=c]{90}{\small 32 views}%
\end{minipage}%
\begin{minipage}[c]{\dimexpr\linewidth-\qlw\relax}%
\qimg{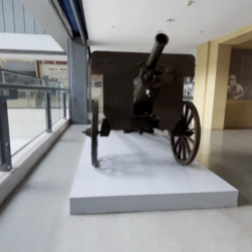}\hfill%
\qimg{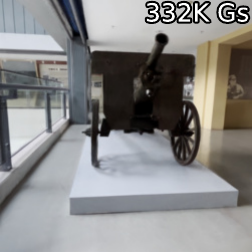}\hfill%
\qimg{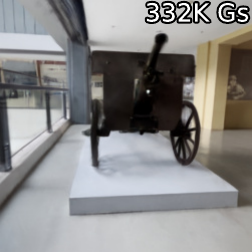}\hfill%
\qimg{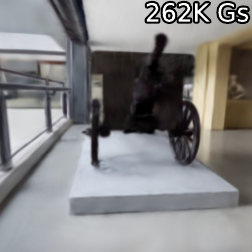}\hfill%
\qimg{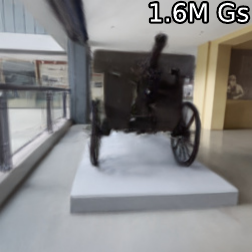}\hfill%
\qimg{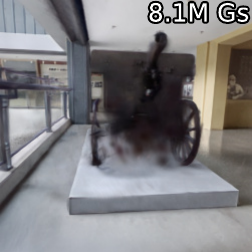}%
\\[2pt]
\qimg{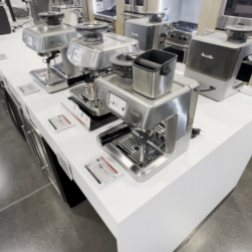}\hfill%
\qimg{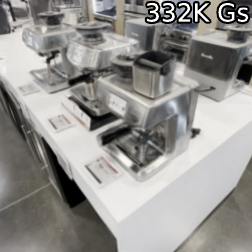}\hfill%
\qimg{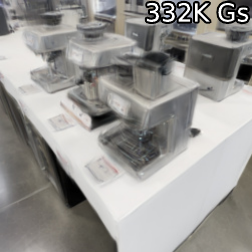}\hfill%
\qimg{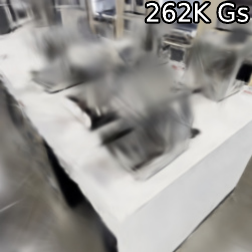}\hfill%
\qimg{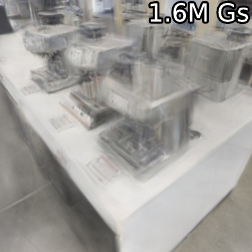}\hfill%
\qimg{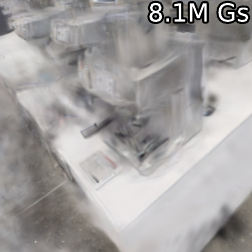}%
\end{minipage}%
\\[2pt]
\begin{minipage}[c]{\qlw}%
\rotatebox[origin=c]{90}{\small 64 views}%
\end{minipage}%
\begin{minipage}[c]{\dimexpr\linewidth-\qlw\relax}%
\qimg{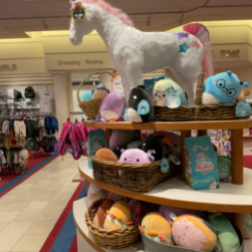}\hfill%
\qimg{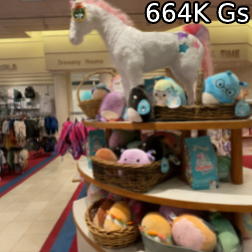}\hfill%
\qimg{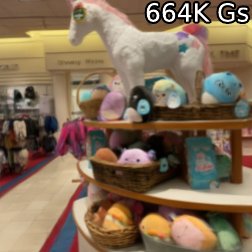}\hfill%
\qimg{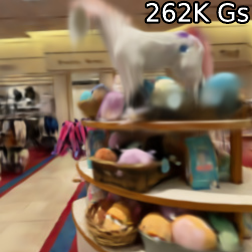}\hfill%
\qimg{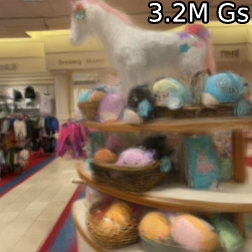}\hfill%
\qimg{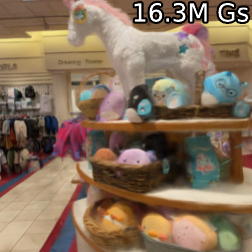}%
\\[2pt]
\qimg{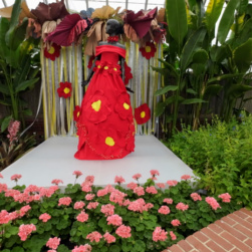}\hfill%
\qimg{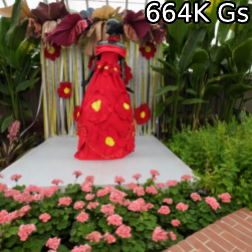}\hfill%
\qimg{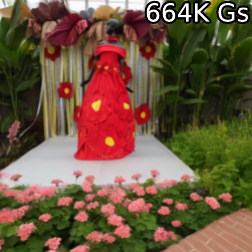}\hfill%
\qimg{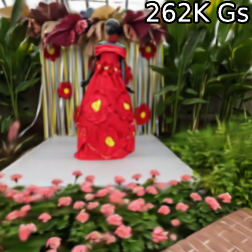}\hfill%
\qimg{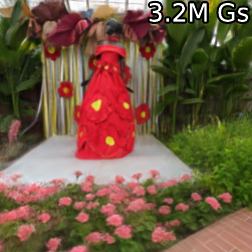}\hfill%
\qimg{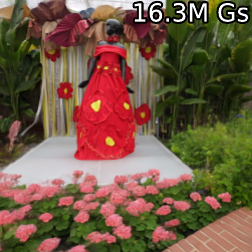}%
\end{minipage}%
\\[2pt]
\begin{minipage}[c]{\qlw}%
\rotatebox[origin=c]{90}{\small 128 views}%
\end{minipage}%
\begin{minipage}[c]{\dimexpr\linewidth-\qlw\relax}%
\qimg{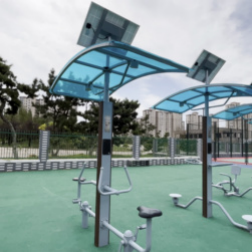}\hfill%
\qimg{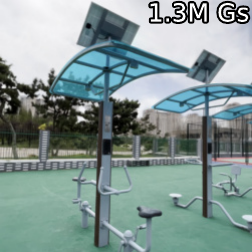}\hfill%
\qimg{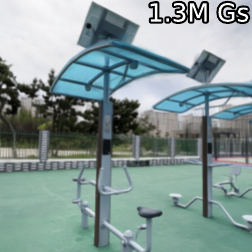}\hfill%
\qimg{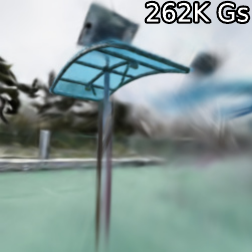}\hfill%
\qimg{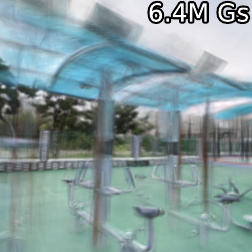}\hfill%
\qimg{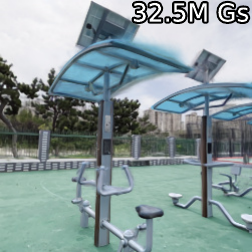}%
\\[2pt]
\qimg{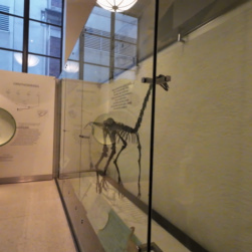}\hfill%
\qimg{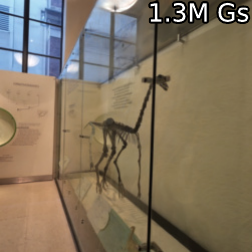}\hfill%
\qimg{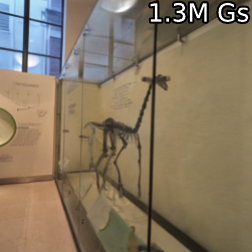}\hfill%
\qimg{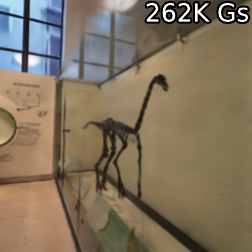}\hfill%
\qimg{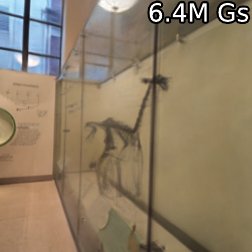}\hfill%
\qimg{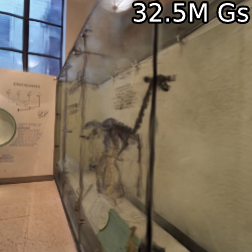}%
\end{minipage}%
\caption{
\textbf{Additional qualitative comparisons on DL3DV} (16 to 128 input views).
}
\label{fig:supp-qual-dl3dv-more}
\end{figure*}

\begin{figure*}[t]
\centering
\ifdefined\sqrcw\else\newlength{\sqrcw}\fi\setlength{\sqrcw}{\dimexpr(\linewidth-16pt)/11\relax}%
\ifdefined\sqriw\else\newlength{\sqriw}\fi\setlength{\sqriw}{\dimexpr2\sqrcw+1pt\relax}%
\newcommand{\sqrctx}[1]{\includegraphics[width=\sqrcw]{figures/supp_qual_re10k/assets/#1}}%
\newcommand{\sqrimg}[1]{\includegraphics[width=\sqriw]{figures/supp_qual_re10k/assets/#1}}%
\newcommand{\sqrscene}[1]{%
\begin{minipage}[c]{\dimexpr3\sqrcw+2pt\relax}%
\setlength{\lineskip}{1pt}\setlength{\lineskiplimit}{1pt}%
\sqrctx{#1_ctx0.pdf}\hspace{1pt}\sqrctx{#1_ctx1.pdf}\hspace{1pt}\sqrctx{#1_ctx2.pdf}\\%
\sqrctx{#1_ctx3.pdf}\hspace{1pt}\sqrctx{#1_ctx4.pdf}\hspace{1pt}\sqrctx{#1_ctx5.pdf}%
\end{minipage}\hspace{2pt}%
\begin{minipage}[c]{\sqriw}\sqrimg{#1_gt.pdf}\end{minipage}\hspace{2pt}%
\begin{minipage}[c]{\sqriw}\sqrimg{#1_ours.pdf}\end{minipage}\hspace{2pt}%
\begin{minipage}[c]{\sqriw}\sqrimg{#1_yonosplat.pdf}\end{minipage}\hspace{2pt}%
\begin{minipage}[c]{\sqriw}\sqrimg{#1_tokengs.pdf}\end{minipage}%
}
\makebox[\dimexpr3\sqrcw+2pt\relax]{\small Context}\hfill%
\makebox[\sqriw]{\small GT}\hfill%
\makebox[\sqriw]{\small Ours}\hfill%
\makebox[\sqriw]{\small YoNoSplat}\hfill%
\makebox[\sqriw]{\small TokenGS}%
\\[1pt]
\sqrscene{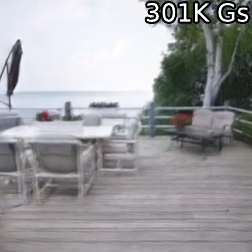}\\[4pt]
\sqrscene{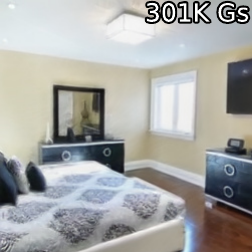}\\[4pt]
\sqrscene{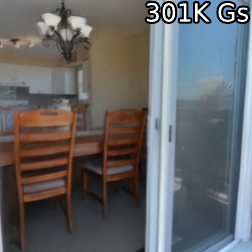}\\[4pt]
\sqrscene{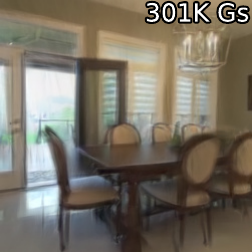}\\[4pt]
\sqrscene{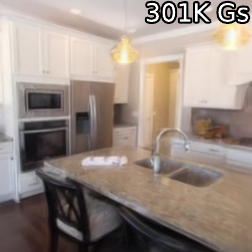}\\[4pt]
\sqrscene{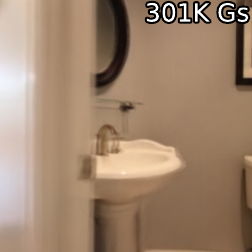}\\[4pt]
\sqrscene{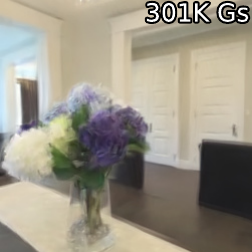}%
\caption{\textbf{Additional qualitative comparisons on RealEstate10K} (6 input views).
}
\label{fig:supp-qual-re10k}
\end{figure*}

\fi

\end{document}